%% file: main.tex
\documentclass[10pt,twocolumn,letterpaper]{article}

\usepackage[pagenumbers]{cvpr} % To force page numbers, e.g. for an arXiv version


\definecolor{cvprblue}{rgb}{0.21,0.49,0.74}
\usepackage[pagebackref,breaklinks,colorlinks,allcolors=cvprblue]{hyperref}

\usepackage{url}            % simple URL typesetting
\usepackage{booktabs}       % professional-quality tables
\usepackage{amsfonts}       % blackboard math symbols
\usepackage{nicefrac}       % compact symbols for 1/2, etc.
\usepackage{microtype}      % microtypography
\usepackage{xcolor}         % colors
\usepackage{xspace}
\usepackage{graphicx}
\usepackage{tabu}
\usepackage{multirow}
\usepackage{pifont}
\usepackage{amsmath}
\usepackage{listings}
\usepackage[frozencache,cachedir=.]{minted}

\def\paperID{268} % *** Enter the Paper ID here
\def\confName{CVPR}
\def\confYear{2025}

\title{HRSAM: Efficient Interactive Segmentation in High-Resolution Images}

\author{%
    You Huang, Wenbin Lai, Jiayi Ji, Liujuan Cao\thanks{Corresponding author}, Shengchuan Zhang, Rongrong Ji \\
    Key Laboratory of Multimedia Trusted Perception and Efficient Computing, \\
    Ministry of Education of China, Xiamen University \\
    {\tt\small youhuang0607@gmail.com, 23020231154143@stu.xmu.edu.cn, jjyxmu@gmail.com}\\
    {\tt\small caoliujuan@xmu.edu.cn, zsc\_2016@xmu.edu.cn, rrji@xmu.edu.cn}\\
}

\begin{document}
\maketitle
\input{sec/0_abstract}    
\input{sec/1_intro}
\input{sec/2_related_work}

\input{sec/3_method}
\input{sec/4_experiments}
\input{sec/5_conclusion}
{
    \small
    \bibliographystyle{ieeenat_fullname}
    \bibliography{main}
}

\input{sec/X_suppl}

\end{document}

%% file: sec/0_abstract.tex
\begin{abstract}
The Segment Anything Model (SAM) has advanced interactive segmentation but is limited by the high computational cost on high-resolution images. This requires downsampling to meet GPU constraints, sacrificing the fine-grained details needed for high-precision interactive segmentation. To address SAM's limitations, we focus on visual length extrapolation and propose a lightweight model named HRSAM. The extrapolation enables HRSAM trained on low resolutions to generalize to high resolutions. We begin by finding the link between the extrapolation and attention scores, which leads us to base HRSAM on Swin attention. We then introduce the Flexible Local Attention (FLA) framework, using CUDA-optimized Efficient Memory Attention to accelerate HRSAM. Within FLA, we implement Flash Swin attention, achieving over a $35\%$ speedup compared to traditional Swin attention, and propose a KV-only padding mechanism to enhance extrapolation. We also develop the Cycle-scan module that uses State Space models to efficiently expand HRSAM's receptive field. We further develop the HRSAM++ within FLA by adding an anchor map, providing multi-scale data augmentation for the extrapolation and a larger receptive field at slight computational cost. Experiments show that, under standard training, HRSAMs surpass the previous SOTA with only $38\%$ of the latency. With SAM-distillation, the extrapolation enables HRSAMs to outperform the teacher model at lower latency. Further finetuning achieves performance significantly exceeding the previous SOTA. Code is available at \href{https://github.com/YouHuang67/High-Resolution-Segment-Anything.git}{https://github.com/YouHuang67/High-Resolution-Segment-Anything.git}
\end{abstract}

%% file: sec/1_intro.tex
\section{Introduction}
\label{sec:intro}

The Segment Anything Model (SAM)~\cite{kirillov2023segany} stands as a cornerstone in interactive segmentation~\cite{DavidAcuna2018EfficientIA,ShiyinZhang2020InteractiveOS,KonstantinSofiiuk2021RevivingIT,XiChen2022FocalClickTP,ZhengLinFocusCutDI,QinLiu2022PseudoClickII} and extends its utility to various downstream tasks in computer vision~\cite{ma2023segment,mazurowski2023segment,lai2023lisa,wu2023medical,yu2023inpaint,wang2023review}. As an interactive segmentation model, SAM takes simple manual inputs like clicks~\cite{huang2023interformer}, bounding boxes~\cite{kirillov2023segany} and coarse masks~\cite{KonstantinSofiiuk2021RevivingIT} to predict precise segmentation results, thus reducing the manual labeling costs associated with image segmentation. This study focuses on mainstream click-based interactive segmentation and further enhances SAM's performance.

SAM utilizes large Vision Transformers (ViTs)~\cite{AlexeyDosovitskiy2020AnII, YanghaoLiExploringPV} as an encoder to process images, but the implementation is limited to handling fixed input resolutions of $1024 \times 1024$. 
In real-world applications, interactive segmentation models often need to handle high-resolution images (exceeding $4096 \times 4096$) to capture the rich details essential for precise segmentation on datasets like HQSeg44K~\cite{sam_hq}. SAM and the variants~\cite{sam_hq,liu2024rethinking} are constrained by computational limits on commonly used GPUs, forcing downsampling of high-resolution images to lower resolutions (\ie $1024^2$\protect\footnote{In this paper, any resolution expressed as $N^2$ (\eg $1024^2$, $2048^2$, $3072^2$, $4096^2$) refers to an image resolution of $N \times N$.}), which sacrifices detail and impacts segmentation accuracy.
Thus, this work aims at developing a lightweight SAM-distilled ViT that efficiently processes high-resolution images.
Specifically, we focus on the \emph{visual length extrapolation}~\cite{song2024ba,leroy2023win}, which enables a ViT distilled from SAM on low-resolution images ($1024^2$) to generalize effectively to high-resolution input images (\eg $2048^2$) without high-resolution training.

\begin{figure*}[!t]
    \centering
    \vspace{-1em}
    \includegraphics[width=0.9\textwidth]{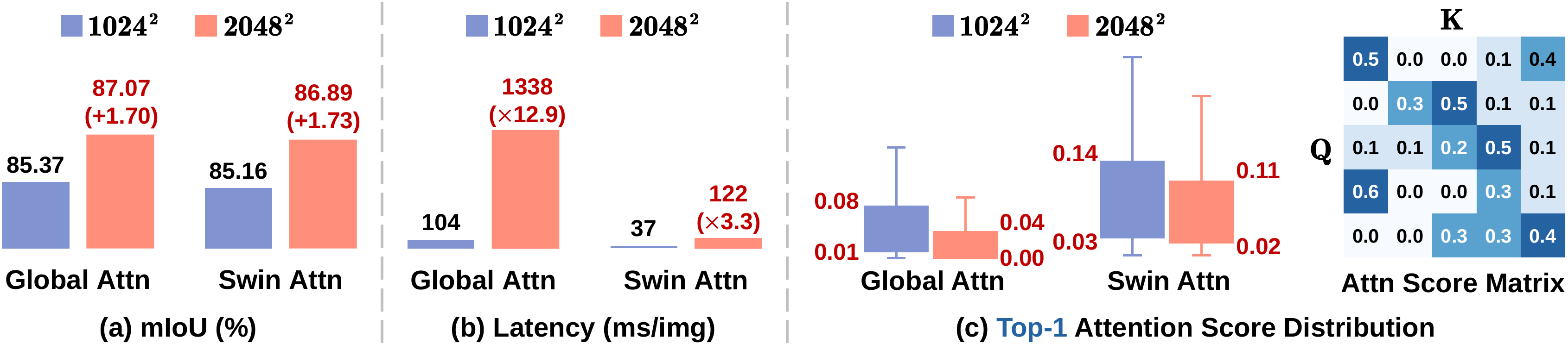}
    \vspace{-.4em}
    \caption{Analysis of visual length extrapolation over Global-attn and Swin-attn ViTs.
    (a) illustrates the mean IoU at each interaction step under the standard testing protocol of interactive segmentation on HQSeg-44K for both ViTs, with input resolutions of $1024^2$ and $2048^2$. (b) shows the inference latency per image for both ViTs at different input resolutions. (c) presents a box plot of the top-1 attention scores (post-softmax) for each image token across all attention computations in both ViTs, with the top $10\%$ values removed for visual clarity and red labels indicating the 0.25 and 0.75 quantiles. Global-attn ViT shows a more pronounced reduction in these scores at higher resolutions.}
    \label{fig:intro-overview}
    \vspace{-1em}
\end{figure*}

We start by exploring the link between visual length extrapolation and attention computations. We build a light ViT with traditional global attention~\cite{AshishVaswani2017AttentionIA} (as in SAM), implemented by modern techniques~\cite{touvron2023llama, fang2023eva} including rotary position embeddings (RoPE)~\cite{su2024roformer} and Efficient Memory Attention (EMA)~\cite{DBLP:journals/corr/abs-2112-05682,xformers_ops} to reduce inference memory. We refer to this model as Global-attn ViT that is trained to align its image embeddings with those of SAM encoder (as a teacher in the distillation) using a simple MSE loss.
However, we find that Global-attn ViT suffers from suboptimal balance between computational efficiency and performance at $2048^2$ images. As shown in Figure~\ref{fig:intro-overview} (a)(b), increasing the input image size from $1024^2$ to $2048^2$ increases computation time by $12.9\times$, while the mIoU only improves by $1.7\%$. Figure~\ref{fig:intro-overview} (c) shows the top-1 post-softmax attention scores for each image token in Global-attn ViT, revealing a significant decrease in these scores as resolution increases. This suggests a reduced focus on critical tokens in high-resolution images, potentially limiting the ability of visual length extrapolation.

The shift in attention score distribution results from changes in each token’s receptive field as input resolution increases. Then, we replace global attention with traditional shifted window (Swin) attention~\cite{liu2021swin,liu2022swin} to fix each token’s perception range, and train the model equivalently, resulting in Swin-attn ViT. As shown in Figure~\ref{fig:intro-overview} (c), larger image sizes have minimal impact on Swin-attn ViT’s top-1 attention scores. Meanwhile, Swin-attn ViT achieves the same performance gains with significantly lower computational costs on larger input images. Thus, we base our model on Swin-attn ViT. However, there remain three issues with the traditional Swin attention~\cite{liu2021swin,liu2022swin}. 
First, the shifted window’s cyclic-shifting masking mechanism introduces an attention bias that requires $O(N^2)$ space complexity, making it incompatible with Efficient Memory Attention (EMA)~\cite{DBLP:journals/corr/abs-2112-05682,xformers_ops} for acceleration.
Second, window sizes vary at boundary windows under the shift mechanism, weakening the length extrapolation. Finally, window attention restricts each token’s receptive field, reducing global information exchange.

To address these issues, we introduce the Flexible Local Attention (FLA) framework that reorders tokens to align with EMA’s block-diagonal attention mechanism~\cite{xformers_ops}, which accelerates general local attention, \eg attention over windowed or variably partitioned tokens. Then, we propose HRSAM under FLA with three improvements. First, we develop Flash Swin under FLA to remove the attention bias in traditional Swin attention, allowing the use of EMA~\cite{DBLP:journals/corr/abs-2112-05682} for acceleration. Flash Swin achieves equivalent computations with $35\%$ speedup on a $16^2$ window and $52\%$ on a $32^2$ window. Second, HRSAM employs a KV-only padding mechanism with learnable padding tokens under FLA to ensure a consistent number of keys for each token during attention, which enhances visual length extrapolation with ignorable extra computation. Finally, HRSAM incorporates a newly proposed Cycle-scan module for efficient global information exchange. Cycle-scan repeatedly applies State Space Models (SSMs)~\cite{gu2023mamba,liu2024vmamba,zhu2024vision} to scan image token sequences with linear computational complexity, achieving further performance gains at low computational cost.

Under the FLA framework, HRSAM is further extended with an anchor-map-based multi-scale strategy~\cite{shi2024we}, forming HRSAM++. The strategy resizes any input image (e.g., $1024^2$, $2048^2$) to a fixed anchor map of $512^2$ as a supplementary input and share the weights across the differently scaled inputs, with multi-scale fusion via the proposed Cycle-scan. HRSAM++ provides two key advantages over HRSAM: enhanced multi-scale data augmentation and an expanded token receptive field. Consequently, the fixed $512^2$ anchor map adds small extra costs but yields notable performance gains.

We conduct experiments with three training settings. First, under fair data conditions, we train HRSAMs on the combined dataset of COCO~\cite{TsungYiLin2014MicrosoftCC}, LVIS~\cite{AgrimGupta2019LVISAD} and HQSeg44K~\cite{sam_hq}, achieving SOTA performance on high-precision interactive segmentation benchmarks~\cite{FedericoPerazzi2016ABD, sam_hq} with only $38\%$ of the previous SOTA's latency. Second, we perform simple distillation from SAM-ViT-Huge, where the $1024^2$-trained HRSAMs effectively scale to $2048^2$, surpassing the teacher in both performance and computational efficiency. Finally, finetuning the distilled HRSAMs on HQSeg44K~\cite{sam_hq} yields performance that significantly surpasses previous SOTA methods~\cite{liu2024rethinking}.

Our main contributions are as follows:
\begin{itemize}
    \item We establish the link between ViTs' visual length extrapolation capability and their top-1 attention scores, which guides the design of our proposed HRSAM's base model.
    \item HRSAM is further implemented under the proposed FLA framework, consisting of the newly developed Flash Swin, KV-only padding and Cycle-scan module.
    \item We further propose HRSAM++ using an anchor-map-based multi-scale strategy under FLA to achieve significant performance gains with slight extra computation. 
    \item Experimental results demonstrate HRSAMs' effectiveness, under SAM-distilled training and fair data conditions, outperforming the previous SOTA at only $38\%$ of the latency.
\end{itemize}

%% file: sec/2_related_work.tex
\begin{figure*}[!ht]
    \centering
    \includegraphics[width=0.88\textwidth]{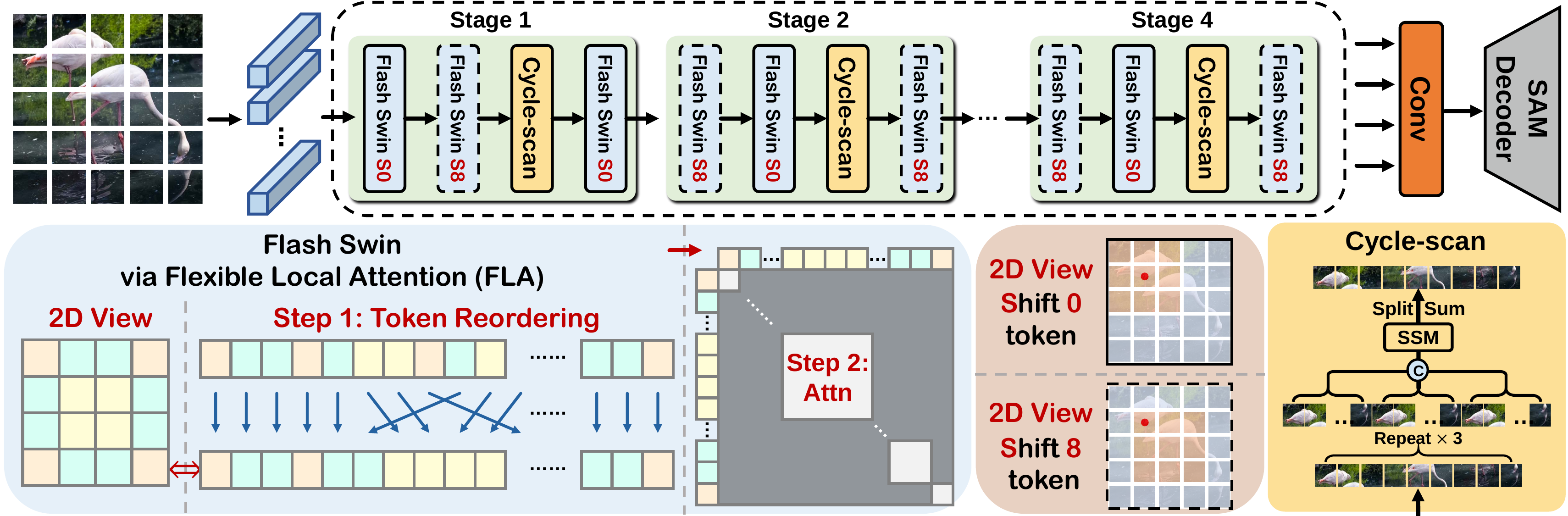}
    \vspace{-0.5em}
    \caption{
    Overview of our proposed HRSAM. HRSAM contains four stages, each with three Flash Swin modules and one Cycle-scan module. Alternating Flash Swin modules apply shifts of 0 and 8 tokens. Each Flash Swin is implemented within the FLA framework, incorporating token reordering (via index mapping) and local attention computation accelerated by EMA. Outputs from all stages are fused through summation and refined via a convolutional block to create final image embeddings fed into the SAM decoder.
    }
    \label{fig:method-overview}
    \vspace{-.5em}
\end{figure*}

\section{Related Work}
\label{sec:relatedwork}

\noindent\textbf{Interactive Segmentation.}
The integration of deep networks into interactive segmentation begins with DIOS~\cite{NingXu2016DeepIO}, leading to advancements in click-based methods~\cite{KevisKokitsiManinis2017DeepEC,ZhuwenLi2018InteractiveIS,ZhengLin2020InteractiveIS,JangWonDong2019InteractiveIS,KonstantinSofiiuk2020fBRSRB}. Subsequent methods focus on enhancing various aspects of interactive segmentation~\cite{DavidAcuna2018EfficientIA,ShiyinZhang2020InteractiveOS,KonstantinSofiiuk2021RevivingIT,XiChen2022FocalClickTP,ZhengLinFocusCutDI,QinLiu2022PseudoClickII,QinLiu2022SimpleClickII,huang2023interformer,xu2024structured,zhang2024leveraging,li2023interactive,yan2023piclick,zhou2023interactive,rana2023dynamite,li2023multi}. SAM~\cite{kirillov2023segany} improves the inference latency by reusing image features and leads to various downstream applications~\cite{ma2023segment,mazurowski2023segment,lai2023lisa,wu2023medical,yu2023inpaint,wang2023review,xu2024rap,cheng2023sam,huang2024focsam}. However, SAM struggles with the high-resolution images.

\noindent\textbf{Efficient Attention.}
Attention mechanisms~\cite{AshishVaswani2017AttentionIA} have made significant strides in computer vision~\cite{AlexeyDosovitskiy2020AnII,gu2022multi,khan2022transformers,RobinStrudel2021SegmenterTF,Xie_Wang_Yu_Anandkumar_Alvarez_Luo_2021,yuan2021hrformer,fang2023eva,wang2023internimage}. Meanwhile, the high computational complexity of attention mechanisms leads to extensive research on efficient attention~\cite{liu2021swin,Zhuoran_Mingyuan_Haiyu_Shuai_Hongsheng_2021,Guo_Qiu_Liu_Shao_Xue_Zhang_2019,Xie_Wang_Yu_Anandkumar_Alvarez_Luo_2021,liu2021swin,Vaswani_Ramachandran_Srinivas_Parmar_Hechtman_Shlens_2021,YanghaoLiExploringPV,han2023flatten,Ho_Kalchbrenner_Weissenborn_Salimans_2019,Huang_Wang_Wei_Huang_Shi_Liu_Huang_2020,Chu_Tian_Wang_Zhang_Ren_Wei_Xia_Shen_2021,Wang_Yao_Chen_Cai_He_Liu_2021,Yang_Li_Zhang_Dai_Xiao_Yuan_Gao_Redmond_Cloud_Ai,Dong_Bao_Chen_Zhang_Yu_Yuan_Chen_Guo_2022,Ren_Li_Wang_Xiao_Chang}. Flash Attention~\cite{dao2022flashattention,dao2023flashattention2} and EMA~\cite{DBLP:journals/corr/abs-2112-05682} re-implement the vanilla attention mechanism efficiently at the CUDA level, reducing the quadratic space complexity to linear and accelerating computations. This study explores Swin attention to enhance ViT on high-resolution images.

\noindent\textbf{Visual Length Extrapolation.} 
Length extrapolation refers to ViTs' ability to generalize to longer inputs than those they are trained on~\cite{song2024ba}, which has been explored in NLP~\cite{press2021train,chi2022kerple}. In vision, enhancing length extrapolation involves modifications of ViTs, \eg adjustments to positional embeddings and patch sizes~\cite{lee2023pix2struct,beyer2023flexivit,8,18,63,34_flex,20,31_felex,yin2021adavit}, using sequence packing~\cite{dehghani2023patch}, or masking most of image tokens~\cite{leroy2023win}. Recent efforts adapt post-training attention computations to handle scaled input images~\cite{song2024ba}. We employ the Swin attention to achieve the visual length extrapolation.

\noindent\textbf{Visual State Space Model.}
Several efforts extend the State-Space Models~\cite{gu2023mamba} to vision, \eg generation~\cite{fu2024mddose, hu2024zigma, shen2024gamba}, multi-modal tasks~\cite{yang2024remamber, li2024spikemba, qiao2024vlmamba, wan2024sigma, zhao2024cobra}, medical image analysis~\cite{ma2024umamba, wang2024mambaunet, ye2024pmamba, liu2024swinumamba, ruan2024vmunet} and remote sensing~\cite{chen2024rsmamba, chen2024changemamba, liu2024rscama}. Various studies~\cite{zhu2024vision,liu2024vmamba,li2024mamband,yang2024plainmamba,huang2024localmamba,long2024dgmamba,pei2024efficientvmamba,patro2024simba} explore 2D scanning strategies in vision tasks. The proposed Cycle-scan simply performs repeated scanning over flattened token sequences for efficient global information exchange.

%% file: sec/3_method.tex
\section{Method}

This study introduces HRSAM to achieve high-precision segmentation on high-resolution images. Section~\ref{subsec:samoverview} reviews the overall structure. Section~\ref{subsec:fla} presents the proposed FLA framework. Section~\ref{subsec:flashswin}~\ref{subsec:kvpadding}
describe Flash Swin and KV-only padding based on FLA. Section~\ref{subsec:cswin} details the Cycle-scan module. Section~\ref{subsec:msfusion} discusses the multi-scale strategy.

\subsection{Overview of HRSAM}
\label{subsec:samoverview}

\noindent\textbf{SAM pipeline.} 
SAM includes an image encoder (ViT), a compact prompt encoder, and a lightweight decoder for efficient segmentation. Before interactive segmentation, images are resized to $1024^2$ and processed by the image encoder to produce reusable embeddings. During interaction, the prompt encoder converts user inputs (\eg clicks, boxes, masks) into prompt embeddings, which the decoder combines with image embeddings to produce segmentation results. The SAM encoder comprises over $95\%$ of parameters and is critical to performance. Thus, our HRSAM focuses on optimizing the encoder without altering other modules.

\noindent\textbf{HRSAM encoder.} 
As illustrated in Figure~\ref{fig:intro-overview}, our HRSAM comprises four stages with 12 blocks similar to ViT-Base~\cite{AlexeyDosovitskiy2020AnII}. Each block consists of the proposed Flash Swin attention module and a traditional FFN layer~\cite{AlexeyDosovitskiy2020AnII}. The 12 blocks are evenly distributed across the four stages, with each stage containing three blocks. To enable global information exchange, the proposed Cycle-scan module is incorporated before the final Flash Swin attention in each stage.
HRSAM aggregates intermediate outputs from each stage rather than only the final stage. These outputs are fused through summation and processed by a convolutional block to produce the final image embeddings. Besides, HRSAM uses Rotary Position Embedding (RoPE)~\cite{su2024roformer} to avoid attention biases in the positional encoding, thereby facilitating the use of EMA~\cite{DBLP:journals/corr/abs-2112-05682} for faster, memory-efficient attention computation.

\noindent\textbf{Efficient Memory Attention.}
EMA~\cite{DBLP:journals/corr/abs-2112-05682} provides a memory-efficient attention algorithm by segmenting computations for parallel processing across GPU CUDA cores, significantly reducing memory usage. 
However, SAM’s attention includes relative positional encoding, requiring $O(N^2)$ space complexity, which is incompatible with EMA.
Instead, HRSAM removes this limitation by replacing SAM's relative positional encoding with RoPE~\cite{su2024roformer} and introducing the FLA framework to enable EMA compatibility with local attention. This design allows HRSAM to efficiently handle attention computations on high-resolution inputs, overcoming memory constraints on standard GPUs like the 3090.
The following section details the proposed FLA framework.

\subsection{Flexible Local Attention Framework}
\label{subsec:fla}

\noindent\textbf{Traditional Attention.}
Consider a sequence of $L = \frac{H}{16} \times \frac{W}{16}$ $C$-dimensional image tokens (\ie embeddings) $\mathbf{X} \in \mathbb{R}^{L \times C}$, derived from an input image of resolution $H \times W$ through a $16 \times 16$ patchification process in a ViT~\cite{dosovitskiy2021image}. The traditional attention performs linear projection over $\mathbf{X}$ to get the $(\mathbf{Q}, \mathbf{K}, \mathbf{V}) = (\mathbf{X}\mathbf{W}_q, \mathbf{X}\mathbf{W}_v, \mathbf{X}\mathbf{W}_v) \in \mathbb{R}^{L \times C}$, and the attention outputs are then computed as follows:
\begingroup
\setlength{\abovedisplayskip}{0.5em}
\setlength{\belowdisplayskip}{0.5em}
\begin{equation}
    \begin{split}
        \text{Attention}(\mathbf{X}) = \text{Softmax}\left(\mathbf{Q}\mathbf{K}^\top
        /
        \sqrt{C}\right)\mathbf{V}.
    \end{split}
    \label{eq:traditionalattention}
\end{equation}
\endgroup
For simplicity, we discuss the single-sample, single-head case in this paper. The multi-sample and multi-head cases can be trivially extended.

\begin{figure}[!ht]
    \vspace{-0.5em}
    \centering
    \includegraphics[width=0.40\textwidth]{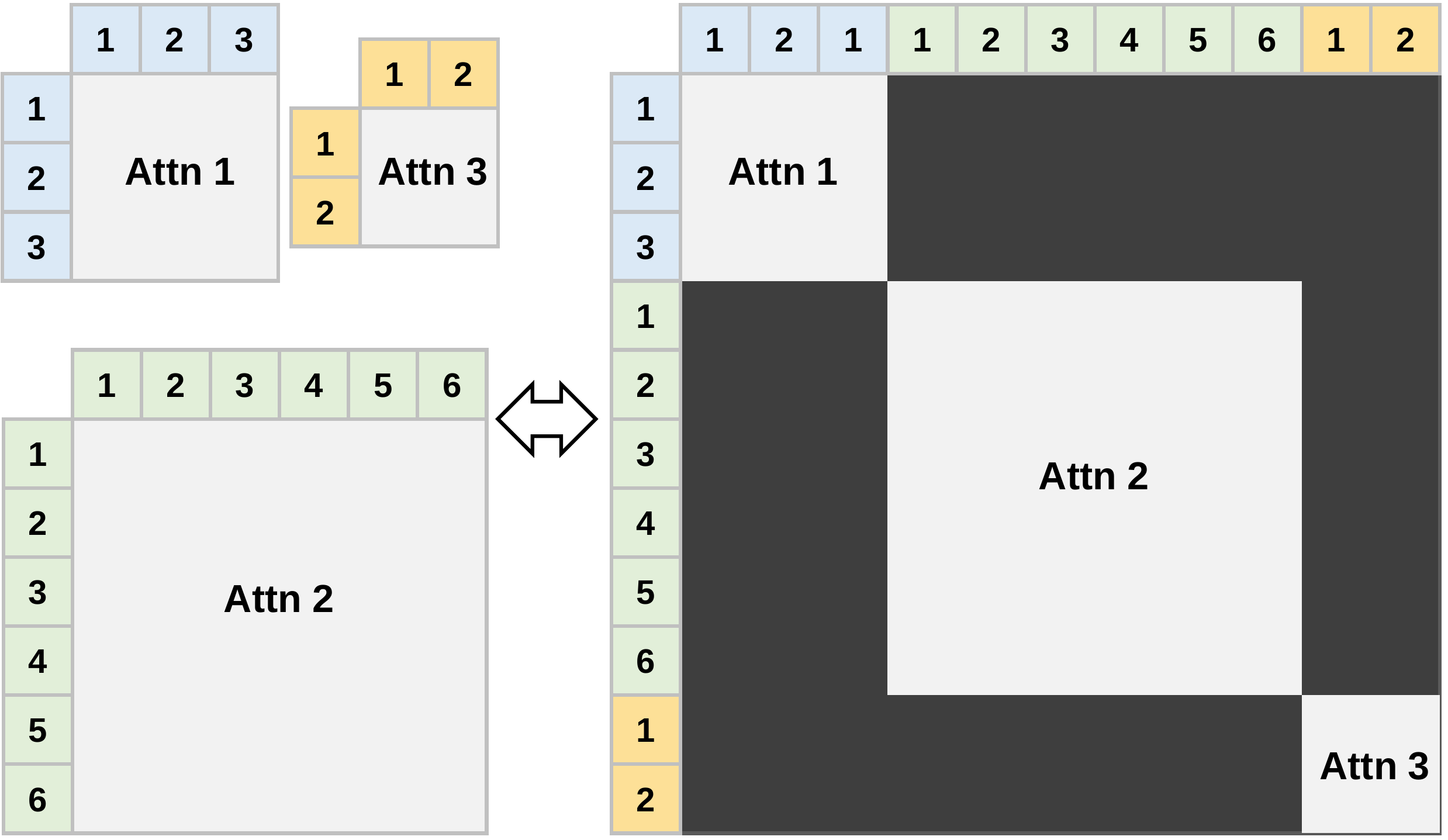}
    \caption{Illustration of EMA's block-diagonal attention~\cite{xformers_ops}. Subsequences of different lengths are shown in blue, yellow and green, with each sequence's attention computation indicated in gray. The black regions represent skipped areas in EMA's computation, optimizing parallelism and reducing memory usage.}
    \label{fig:blockdiagonalattn}
    \vspace{-0.5em}
\end{figure}

\noindent\textbf{Block-diagonal attention.} 
Eq.~\ref{eq:traditionalattention} defines the computation of global attention, where all tokens interact with each other. However, many ViT variants~\cite{liu2021swin,liu2021swin,YanghaoLiExploringPV,han2023flatten,Ho_Kalchbrenner_Weissenborn_Salimans_2019,Huang_Wang_Wei_Huang_Shi_Liu_Huang_2020,Chu_Tian_Wang_Zhang_Ren_Wei_Xia_Shen_2021,Wang_Yao_Chen_Cai_He_Liu_2021,Yang_Li_Zhang_Dai_Xiao_Yuan_Gao_Redmond_Cloud_Ai,Dong_Bao_Chen_Zhang_Yu_Yuan_Chen_Guo_2022,Ren_Li_Wang_Xiao_Chang} employ local attention mechanisms to avoid long-range token interactions, which helps reduce computational complexity. For instance, window attention~\cite{YanghaoLiExploringPV} partitioning the sequence $\mathbf{X}$ into $N$ non-overlapping subsequences $\mathbf{X}^i, 1 \le i \le N$ and the attention is then computed independently for each subsequence:
$ \text{Attention}(\mathbf{X}^i), 1 \le i \le N$.
In implementations, if each subsequence has equal length $|\mathbf{X}^i| = L / N$, the sequences can be transformed into  $\left\{\mathbf{X}^i\right\}_i \in \mathbb{R}^{N \times \frac{L}{N} \times C}$, allowing parallel computation across the subsequences. However, in some cases, the subsequences may have unequal lengths, such as with shift windows. In such scenarios, EMA~\cite{xformers_ops} can be employed to perform block-diagonal attention. For example, consider three subsequences $\mathbf{X}_i, i = 1, 2, 3$, where $|\mathbf{X}^1| = 3$, $|\mathbf{X}^2| = 2$ and $|\mathbf{X}^3| = 6$. As shown in Figure~\ref{fig:blockdiagonalattn}, attention is computed independently for each subsequence (left), which can be equivalently transformed into EMA's block-diagonal attention computation by merging the subsequences and performing attention simultaneously (right). The computation skips black regions at the CUDA level to avoid unnecessary calculations, improving GPU parallel processing efficiency and accelerating local attention.

\noindent\textbf{Flexible Local Attention.}
We use the block-diagonal attention to formulate the flexible local attention. Given a sequence of tokens $\mathbf{X} = \{\mathbf{X}_j\}_{j=1}^{L} \in \mathbb{R}^{L \times C}$, we construct an index mapping: $\mathcal{I}: \{1, 2, \ldots, L\} \rightarrow \{1, 2, \ldots, L\}$ to reorder the tokens in $\mathbf{X}$, which yields
\begingroup
\setlength{\abovedisplayskip}{0.5em}
\setlength{\belowdisplayskip}{0.5em}
\begin{equation}
    \mathbf{X}_\mathcal{I} = \text{Reorder}(\mathbf{X}, \mathcal{I}) = \left({\mathbf{X}_{\mathcal{I}(1)}, \mathbf{X}_{\mathcal{I}(2)}, \ldots, \mathbf{X}_{\mathcal{I}(L)}}\right).
\end{equation}
\endgroup
Then, given the specific local attention computation, we further partition this sequence into $N$ non-overlapping subsequences accordingly, \ie,
\begingroup
\setlength{\abovedisplayskip}{0.5em}
\setlength{\belowdisplayskip}{0.5em}
\begin{equation} 
    \mathbf{X}_{\mathcal{I}} = {\bigcup}_{i=1}^N \mathbf{X}_{\mathcal{I}}^i =
    \mathbf{X}_{\mathcal{I}}^1 \cup \mathbf{X}_{\mathcal{I}}^2 \cup \cdots \cup \mathbf{X}_{\mathcal{I}}^N, 
\end{equation}
\endgroup
where given the length $L_i = \Sigma_{k=1}^i |\mathbf{X}_{\mathcal{I}}^k|, 1 \le i \le N$ and $L = L_N$, defining $\mathbf{X}_{\mathcal{I}}^i$ as
\begingroup
\setlength{\abovedisplayskip}{0.5em}
\setlength{\belowdisplayskip}{0.5em}
\begin{equation}
\begin{split}
    \mathbf{X}_{\mathcal{I}}^1 &= (\mathbf{X}_{\mathcal{I}(1)}, \mathbf{X}_{\mathcal{I}(2)}, \ldots, \mathbf{X}_{\mathcal{I}(L_1)}), \\
    \mathbf{X}_{\mathcal{I}}^2 &= (\mathbf{X}_{\mathcal{I}(L_1 + 1)}, \mathbf{X}_{\mathcal{I}(L_1 + 2)}, \ldots, \mathbf{X}_{\mathcal{I}(L_2)}), \\
    &\cdots \\
    \mathbf{X}_{\mathcal{I}}^N &= (\mathbf{X}_{\mathcal{I}(L_{N - 1} + 1)}, \mathbf{X}_{\mathcal{I}(L_{N - 1} + 2)}, \ldots, \mathbf{X}_{\mathcal{I}(L_N)}).
\end{split}
\end{equation}
\endgroup
We perform the block-diagonal attention over $\cup_{i=1}^N \mathbf{X}_{\mathcal{I}}^i$, as illustrated on the right side of Figure~\ref{fig:blockdiagonalattn}, to obtain the attention outputs $\mathbf{Y}^i = \text{Attention}(\mathbf{X}_{\mathcal{I}}^i), 1 \le i \le N$. Finally, we construct the inverse index mapping $\mathcal{I}^{-1}$ satisfying $\mathcal{I}^{-1}(\mathcal{I}(j)) = j, 1 \le j \le L$ to produce the final results $\text{Reorder}(\cup_{i=1}^N \mathbf{Y}^i, \mathcal{I}^{-1})$. 
In summary, by constructing the index mapping $\mathcal{I}$ and defining the subset partition lengths $\{L_i\}_{i=1}^N$, we can implement a specific FLA.
Next, we leverage this FLA framework to formulate Flash Swin attention.

\subsection{Flash Swin Attention}
\label{subsec:flashswin}

\begin{figure}[!ht]
    \centering
    \vspace{-0.5em}
    \includegraphics[width=0.38\textwidth]{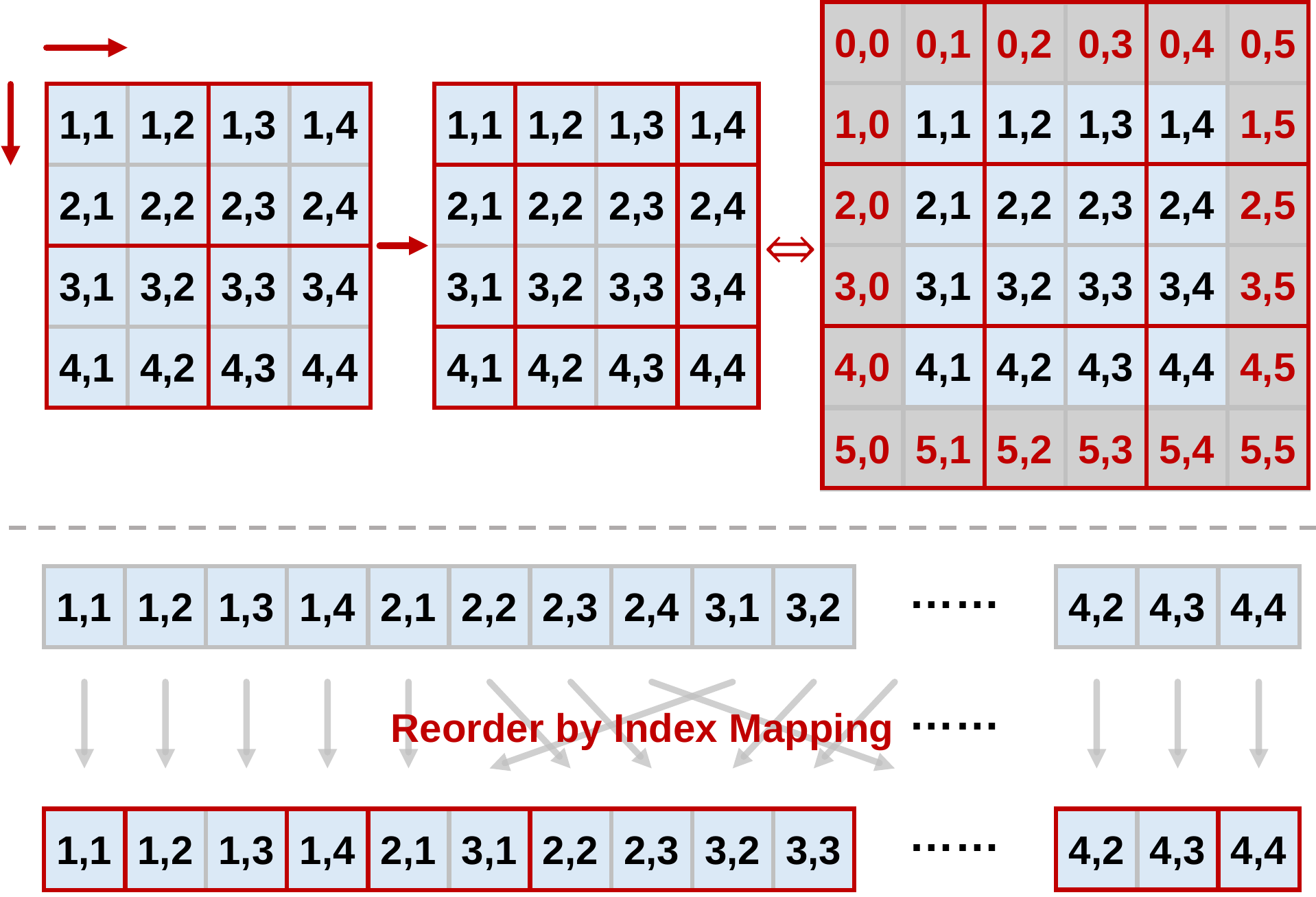}
    \vspace{-0.5em}
    \caption{Illustration of Flash Swin's index mapping for a simple case. The top part shows the shift operation from a 2D perspective, while the bottom part depicts the actual index mapping.}
    \label{fig:flashswin}
    \vspace{-0.5em}
\end{figure}

\noindent\textbf{Non-shift window.}
For a token sequence $\mathbf{X} \in \mathbb{R}^{L\times C}$ of length $L = hw$, where $h \!=\! H / 16 $ and $ w \!=\! W / 16$, we can transform it into a 2D structure through a vanilla coordinate mapping $\mathcal{I}_{\text{van}}: \{1, 2, \ldots, L\} \rightarrow \{1, 2, \ldots, h\} \times \{1, 2, \ldots, w\}$, \ie 
% \begin{equation}
%     \mathcal{I}_{\text{van}}(j; h, w) \!=\! \left(\left\lfloor \frac{j - 1}{w}\right\rfloor, (j - 1) \bmod w\right).
% \end{equation}
\begingroup
\setlength{\abovedisplayskip}{0.5em}
\setlength{\belowdisplayskip}{0.5em}
\begin{equation}
% \small
    \mathcal{I}_{\text{van}}(j; h, w) \!=\! \left(\left\lfloor \frac{j - 1}{w}\right\rfloor \!+\! 1, (j - 1) \bmod w \!+\! 1\right).
\end{equation}
\endgroup
Given the window size $k$ (assume $h,w$ are multiples of the window size $k$), each token is exactly positioned in a window, as shown in the top-left of Figure~\ref{fig:flashswin}. We define the mapping of the coordinates $\mathcal{I}_{\text{win}}: \{1, 2, \ldots, h\} \times \{1, 2, \ldots, w\} \rightarrow \{1, 2, \ldots, L\}$ to ensure that tokens within the same window are arranged in adjacent positions under FLA, as follows:
\begingroup
\setlength{\abovedisplayskip}{0.5em}
\setlength{\belowdisplayskip}{0.5em}
\begin{equation}
\begin{split}
    \mathcal{I}_{\text{win}}(y, x; h, w, k) = \left( \left\lfloor \frac{y - 1}{k} \right\rfloor \cdot \frac{w}{k} + \left\lfloor \frac{x - 1}{k} \right\rfloor \right) k^2 \\
    + \left( (y - 1) \bmod k \right) \cdot k + \left( (x - 1) \bmod k \right) + 1.
\end{split}
\label{eq:nonshift}
\end{equation}
\endgroup
Noting that both mappings are bijective and thus invertible, the non-shift window's index mapping under FLA is given by the composition of the inverse mappings $\mathcal{I}_{\text{val}}^{-1} \circ \mathcal{I}_{\text{win}}^{-1}$. The subset partition length is defined as $L_i = i k^2$.

\noindent\textbf{Shift window.}
For the shift window, we assume a shift of $s \!=\! k / 2$ token units in both the $y$ and $x$ directions, which repositions the windows and results in different window sizes along the boundaries, as shown in the top part of Figure~\ref{fig:flashswin}. Then, we can pad the boundaries with virtual tokens (shown with red coordinates) to maintain consistent window sizes. This requires padding $k - s = k / 2$ virtual tokens in each direction.
Next, we apply the non-shift index mapping to the virtual-token-padded results to obtain the shift index mapping, extending the domains/ranges of $\mathcal{I}_{\text{val}}$ and $\mathcal{I}_{\text{win}}$:
\begingroup
\setlength{\abovedisplayskip}{0.5em}
\setlength{\belowdisplayskip}{0.5em}
\begin{equation}
\begin{split}
    &\{1, 2, \ldots, L\}
    \rightarrow 
    \{1, 2, \ldots, (h + k)(w + k)\}, \\
    &\{1, 2, \ldots, h\}
    \rightarrow 
     \{1 - k / 2, 2 - k / 2, \ldots, h + k / 2\}, \\
    &\{1, 2, \ldots, w\}
    \rightarrow 
     \{1 - k / 2, 2 - k / 2, \ldots, w + k / 2\},
\end{split}
\end{equation}
\endgroup
%For example, $\{1, 2, \ldots, h\}$ is extended to $\{1 - k / 2, 2 - k / 2, \ldots, h + k / 2\}$, 
which leads to the extended mappings:
\begingroup
\setlength{\abovedisplayskip}{0.5em}
\setlength{\belowdisplayskip}{0.5em}
\begin{equation}
\scalebox{0.72}{$
\begin{split}
    &\mathcal{I}_{\text{val}}^\prime(j; h, w, k) = \left(\left\lfloor \frac{j - 1}{w + k}\right\rfloor + 1 - \frac{k}{2}, (j - 1) \bmod (w + k) + 1 - \frac{k}{2}\right),\\
    &\mathcal{I}_{\text{win}}^\prime(y, x; h, w, k) =  \mathcal{I}_{\text{win}}\left(y + \frac{k}{2}, x + \frac{k}{2}; h + k, w + k, k\right).
\end{split}
$}
\end{equation}
\endgroup
The final shift index mapping is derived from the composition $\mathcal{I}_{\text{val}}^{\prime-1} \circ \mathcal{I}_{\text{win}}^{\prime-1}$, omitting virtual token indices to achieve the token reordering shown in the bottom of Figure~\ref{fig:flashswin}.
Appendix~\ref{app:shiftwindow} provides further on constructing this index mapping and calculating subset partition lengths, fully defining the proposed Flash Swin attention within our FLA.

\begin{figure}[!ht]
    \vspace{-0.7em} 
    \centering
    \includegraphics[width=0.40\textwidth]{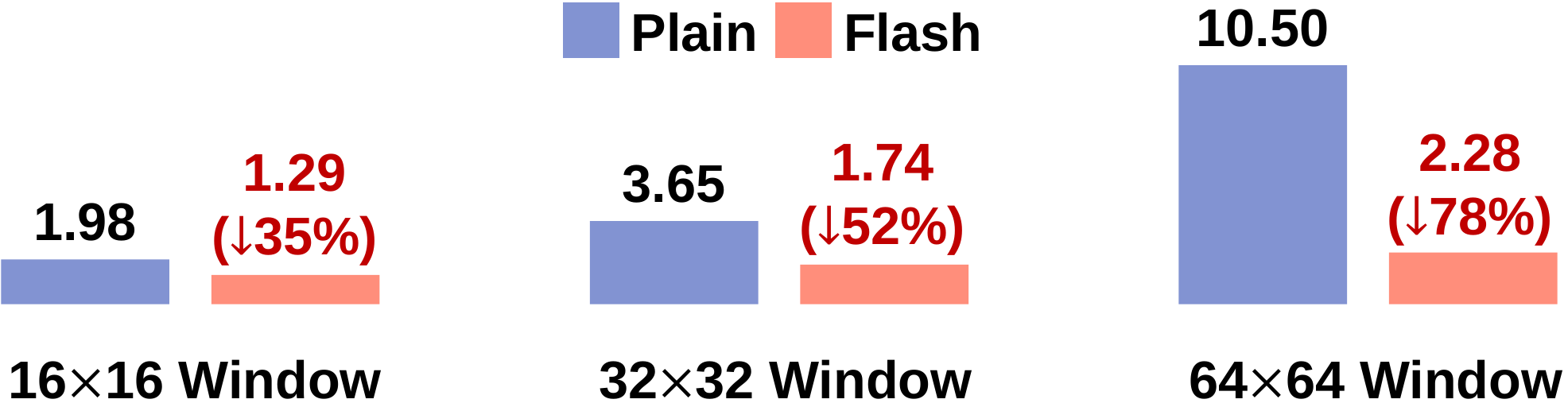}
    \vspace{-0.6em}
    \caption{Latency (ms) of various Swin attention implementations.  The evaluation is conducted with $h = 64$, $w = 64$ and $C = 768$.}
    \label{fig:speedup}
    \vspace{-0.7em}
\end{figure}

\noindent\textbf{Efficiency of Flash Swin.}
Traditional Swin attention~\cite{liu2021swin} requires an attention bias with $O(N^2)$ space complexity to implement cyclic-shifting masking for handling boundary windows' local attention, making it incompatible with EMA. To evaluate efficiency, we compare the traditional (Plain) Swin and our proposed Flash Swin across window sizes $k \!=\! 16, 32, 64$ with shift $s \!=\! k / 2$. As shown in Figure~\ref{fig:speedup}, Flash Swin achieves a significant reduction in latency.

\subsection{KV-only Padding}
\label{subsec:kvpadding}

\noindent\textbf{Extended flexible local attention.}
The block-diagonal attention implemented in EMA~\cite{DBLP:journals/corr/abs-2112-05682} allows $\mathbf{Q}$ and $\mathbf{K}, \mathbf{V}$ to differ in length within each local attention segment, \ie the attention computation on the right side of Figure~\ref{fig:blockdiagonalattn} can have differing numbers of rows and columns, allowing for non-square diagonal blocks. 
Then, we extend the FLA framework by considering differently sized $\mathbf{Q} \in \mathbb{R}^{L_{\text{Q}} \times C}$ and $\mathbf{K}, \mathbf{V} \in \mathbb{R}^{L_{\text{KV}} \times C}$ with $L_{\text{Q}} \neq L_{\text{KV}}$. We partition both $\mathbf{Q}$ and $\mathbf{K}, \mathbf{V}$ into $N$ subsequences, \ie $\{\mathbf{Q}^i\}_{i=1}^N$ and $\{\mathbf{K}^i, \mathbf{V}^i\}_{i=1}^N$. EMA enables efficient local attention between $\{\mathbf{Q}^i\}_{i=1}^N$ and $\{\mathbf{K}^i, \mathbf{V}^i\}_{i=1}^N$, while skipping unnecessary computations.

\begin{figure}[!ht]
    \centering
    \includegraphics[width=0.44\textwidth]{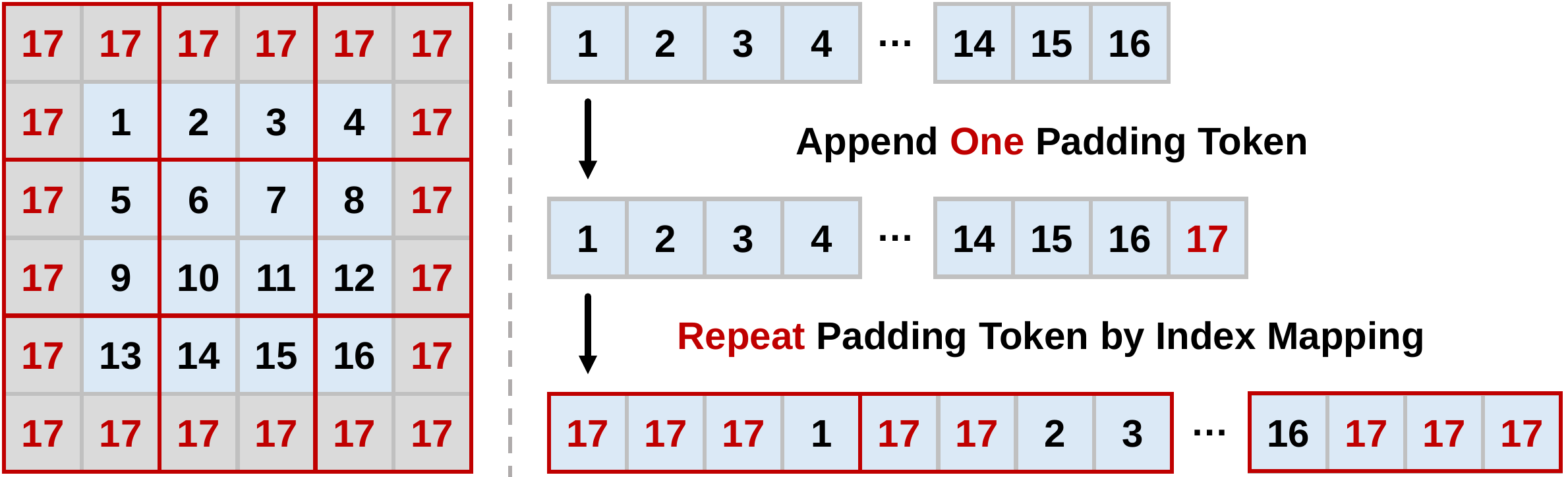}
    \caption{Illustration of KV-only padding by an index mapping for a simple case. The left side shows the 2D perspective of padding, while the right side illustrates the actual index mapping over $\mathbf{K}, \mathbf{V}$.}
    \label{fig:padding}
    \vspace{-0.5em}
\end{figure}

\noindent\textbf{$\mathbf{Q}$ and $\mathbf{K}/\mathbf{V}$'s index mappings.}
We employ the extended FLA framework to make slight modifications to Flash Swin attention. For a given $\mathbf{X} \in \mathbb{R}^{L \times C}$, we compute $\mathbf{Q}, \mathbf{K}, \mathbf{V} \in \mathbb{R}^{L \times C}$ through linear projection. We then apply the previously described shift window index mapping to $\mathbf{Q}$ to form $\{\mathbf{Q}^i\}_{i=1}^N$.
For $\mathbf{K}$ and $\mathbf{V}$, we design an index-mapping-based padding mechanism. As illustrated in Figure~\ref{fig:padding}, we first pad a learnable key/value token at the end of $\mathbf{K}/\mathbf{V}$. We then construct an index mapping, similar to the extended index mapping discussed in the shift window section. Specific details are provided in Appendix~\ref{app:kvpad}. As shown on the left side of Figure~\ref{fig:padding}, this index mapping results in the padding mechanism from a 2D perspective, producing the sets $\{\mathbf{K}^i, \mathbf{V}^i\}_{i=1}^N$ for attention computation under the extended FLA framework.
It’s worth noting that the index mapping applied to $\{\mathbf{Q}^i\}_{i=1}^N$ is bijective, allowing us to reorder the attention results back to the original sequence using its inverse mapping, which is independent of $\mathbf{K}/\mathbf{V}$'s index mapping.

\subsection{Cycle-scan Module}
\label{subsec:cswin}

\noindent\textbf{State Space Models.}
State Space Models (SSMs)~\cite{gu2023mamba} are a class of sequence models designed for capturing long dependencies among sequential tokens, which typically map a one-dimensional sequence $x(t) \in \mathbb{R}$ to $y(t) \in \mathbb{R}$ through an intermediate latent state $h(t) \in \mathbb{R}^N$, formulated as:
\begingroup
\setlength{\abovedisplayskip}{0.5em}
\setlength{\belowdisplayskip}{0.5em}
\begin{equation}
    \begin{split}
        h'(t) = \mathbf{A}h(t) + \mathbf{B}x(t), 
        \quad
        y(t) = \mathbf{C}h(t),\\
    \end{split}
\end{equation}
\endgroup
where the matrices $\mathbf{A} \in \mathbb{R}^{N \times N}$, $\mathbf{B} \in \mathbb{R}^{N \times 1}$ and $\mathbf{C} \in \mathbb{R}^{1 \times N}$ are predefined. Actually, the SSMs are discretized through a zero-order hold rule with a given sample timescale $\Delta \in \mathbb{R}$:
\begingroup
\setlength{\abovedisplayskip}{0.5em}
\setlength{\belowdisplayskip}{0.5em}
\begin{equation}
\small
\begin{split}
    &h_t = \bar{\mathbf{A}}h_{t-1} + \bar{\mathbf{B}}x_t, 
    \quad
    y_t = \bar{\mathbf{C}}h_t, \\
    &\bar{\mathbf{A}} = e^{\Delta \mathbf{A}}, \quad\bar{\mathbf{B}} = (\Delta \mathbf{A})^{-1}\left(e^{\Delta \mathbf{A}} - \mathbf{I} \right)\Delta\mathbf{B}, \quad\bar{\mathbf{C}} = \mathbf{C}. 
\end{split}
\end{equation}
\endgroup
In summary, the SSM operator takes the input sequence $\mathbf{x} = \{x_t\}_{t=1}^L$ as inputs, and produces outputs of the same dimension, $\mathbf{y} = \{y_t\}_{t=1}^L$. SSMs are capable of modeling global information with linear computational complexity. Additionally, SSMs can be readily extended to handle $d$-dimensional sequences $\mathbf{X} \in \mathbb{R}^{d\times L}$, processing each channel independently. In this study, we leverage the recent advancements in selective SSMs, specifically the Mamba~\cite{gu2023mamba}. 

\begin{figure}[!ht]
    \vspace{-0.7em}
    \centering
    \includegraphics[width=0.40\textwidth]{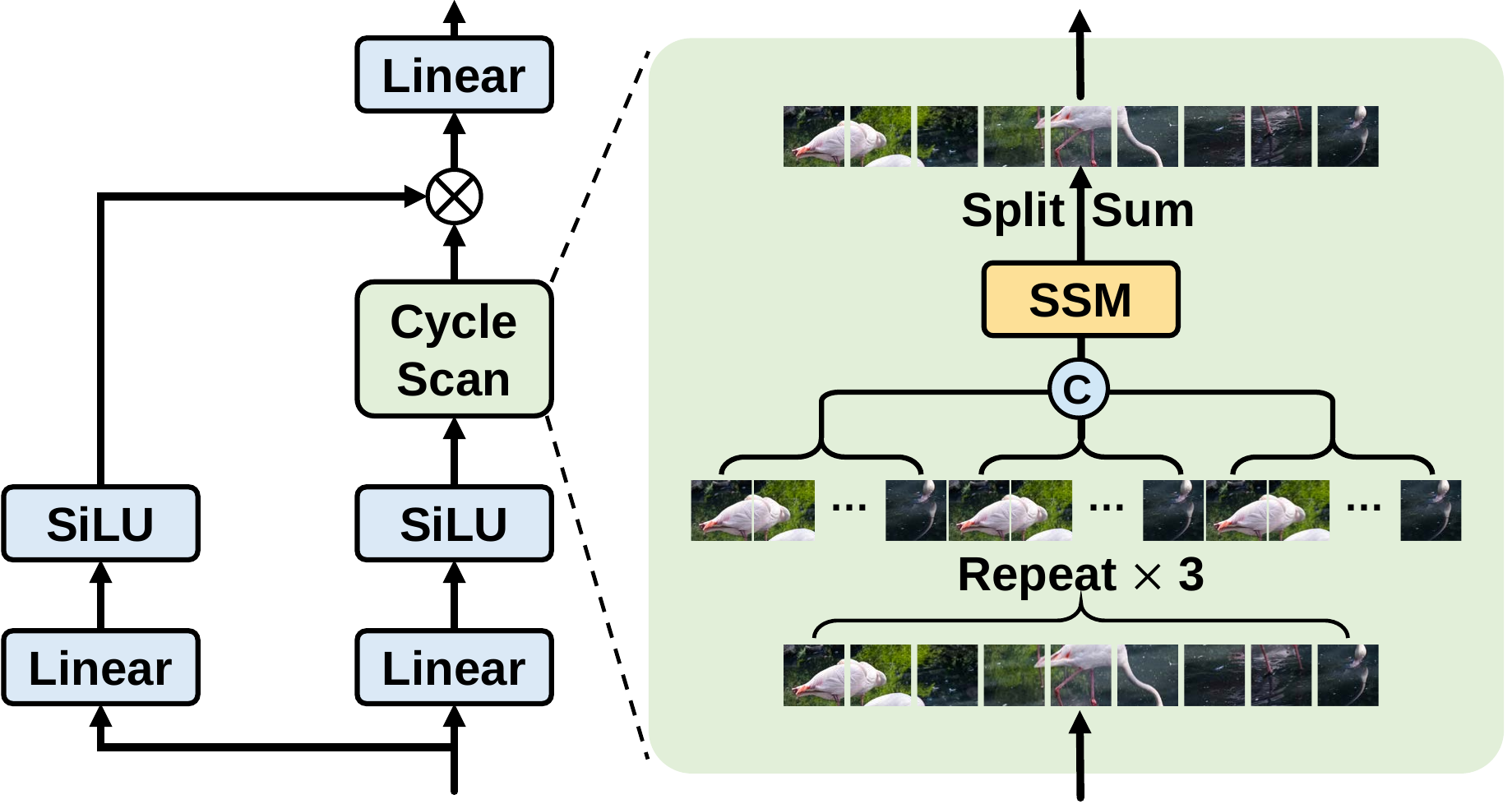}
    \vspace{-0.6em}
    \caption{Illustration of the proposed Cycle-scan module.}
    \label{fig:cyclescan}
    \vspace{-0.7em}
\end{figure}

\noindent\textbf{Cycle-scan strategy.}
Instead of using sophisticated scanning strategies of Mamba SSM in vision~\cite{liu2024vmamba}, we introduce the Cycle-scan strategy that simplifies the scanning process by expanding the image token sequence through replication. As illustrated in Figure~\ref{fig:cyclescan}, the proposed Cycle-scan strategy repeats the token sequence three times and connects them sequentially. The SSM operator then scans the tokens in this order. After scanning, the results are split into the three sequences and are subsequently merged through summation.

\subsection{Multi-scale Fusion}
\label{subsec:msfusion}

\begin{figure}[!ht]
    % \vspace{-0.7em}
    \centering
    \includegraphics[width=0.38\textwidth]{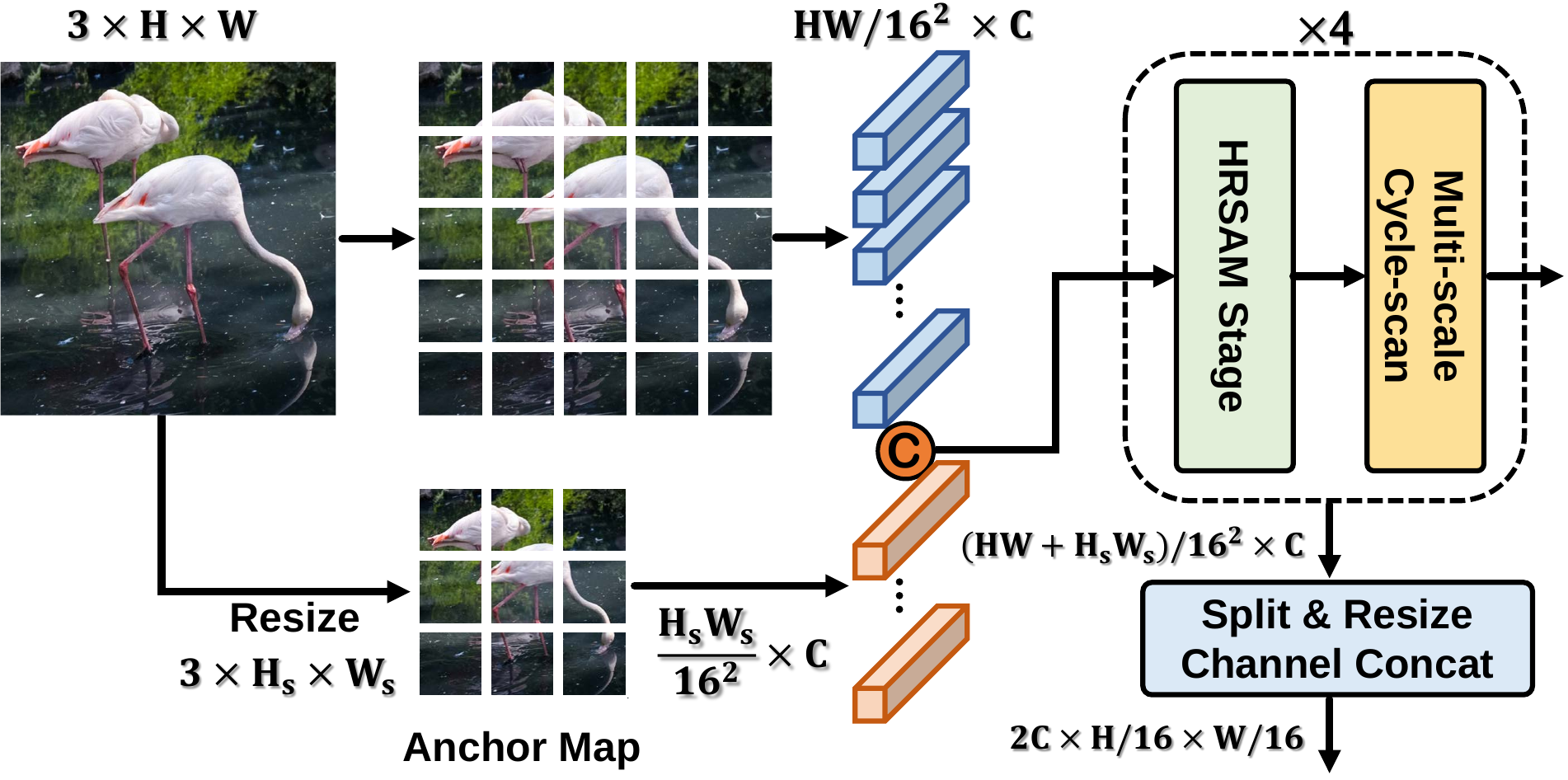}
    \vspace{-0.7em}
    \caption{Overview of HRSAM++. HRSAM++ extends HRSAM with multi-scale inputs at the original size and a smaller anchor map. Each stage’s features are fused by the cycle-scan module, then split, resized, and channel-concatenated to form the outputs.}
    \label{fig:multiscale}
    \vspace{-0.5em}
\end{figure}

\noindent\textbf{Multi-scale inputs.}
As shown in Figure~\ref{fig:multiscale}, HRSAM++ enhances HRSAM by adding a downsampled anchor map of fixed size $H_s \times W_s$ (\eg $512\times512$). Both inputs share the same HRSAM weights, and their outputs are fused at each stage using an extra multi-scale Cycle-scan module.
This design has two benefits. First, the anchor map serves as multi-scale data augmentation, boosting HRSAM’s capacity to handle images at different scales and improving visual length extrapolation. Second, the anchor map provides a larger receptive field to capture more information. Moreover, adding the anchor map introduces slight computational cost.

\noindent\textbf{Multi-scale attention.}
For multi-scale input $\mathbf{X} \in \mathbb{R}^{L \times C}$, where $L \!=\! HW/16^2 + H_sW_s/16^2$, we use a segmented index mapping under FLA. The first $HW/16^2$ tokens follow the standard shift window index mapping, while the remaining $H_sW_s/16^2$ tokens use an additional  shift mapping. Specifically, we define $\mathcal{I}_s(j) \!=\! j - HW/16^2$ and use the compositions $\mathcal{I}_{\text{van}} \circ \mathcal{I}_s$ and $\mathcal{I}_s^{-1} \circ \mathcal{I}_{\text{win}}$ to replace $\mathcal{I}_{\text{van}}$ and $\mathcal{I}_{\text{win}}$ defined in Section~\ref{subsec:flashswin}.
Finally, the segmented index mappings are unified (see Appendix~\ref{app:multiscale} for details), enabling FLA to seamlessly extend to multi-scale inputs while maintaining high parallelism in EMA.

%% file: sec/4_experiments.tex
\section{Experiments}
Section~\ref{subsec:experimentalsetting} details the experimental setup. Section~\ref{subsec:mainresult} discusses the main results. Section~\ref{subsec:scaleanlysis} evaluate HRSAM's scalability to input sizes. Section~\ref{sec:ablationstudy} provides each module's ablation study. Section~\ref{subsec:qualitativeresults} presents the qualitative results.

\begin{table*}[!t]
  \centering
  \resizebox{\textwidth}{!}{
  \begin{tabu}[c]{l c c c c c c c c}
    \toprule
    \multirow{2}{*}{\textbf{Model}} & 
    \multirow{2}{*}{\textbf{Training Data}} &
    \textbf{Inference Time} &
    \multicolumn{3}{c}{\textbf{HQSeg44K} {\small$_{\textbf{Max H/W} > 4000}$}} & 
    \multicolumn{3}{c}{\textbf{DAVIS} {\small$_{\textbf{Max H/W} < 1000}$}} \\
    \cmidrule(lr){4-6} \cmidrule(lr){7-9}
    & & \textbf{20-SPC (ms) $\downarrow$} & 
    5-mIoU $\uparrow$ & NoC90 $\downarrow$ & NoC95 $\downarrow$ &
    5-mIoU $\uparrow$ & NoC90 $\downarrow$ & NoC95 $\downarrow$ \\
    \midrule
    RITM-HRNet32 $_{400}$~\cite{KonstantinSofiiuk2021RevivingIT} & COCO+LVIS & 46 &
    77.72 & 10.01 & 14.58 & 89.75 & 5.34 & 11.45 \\
    FocalClick-SegF-B3-S2 $_{256}$~\cite{XiChen2022FocalClickTP} & COCO+LVIS & 24 &
    84.63 & 8.12 & 12.63 & 90.82 & 5.17 & 11.42 \\
    FocalClick-SegF-B3-S2 $_{384}$~\cite{XiChen2022FocalClickTP} & COCO+LVIS & 26 &
    85.45 & 7.03  & 10.74 & 91.22 & 4.90 & 10.40 \\    
    SimpleClick-ViT-B $_{448}$~\cite{QinLiu2022SimpleClickII} & COCO+LVIS & 31 &
    85.11 & 7.47  & 12.39 & 90.73 & 5.06 & 10.37 \\    
    InterFormer-ViT-B $_{1024}$~\cite{huang2023interformer} & COCO+LVIS & 13 &
    82.62 & 7.17  & 10.77 & 87.79 & 5.45 & 11.88 \\
    SAM-ViT-B $_{1024}$~\cite{kirillov2023segany} & SA-1B & 12 &
    86.16 & 7.46 & 12.42 & 90.95 & 5.14 & 10.74 \\
    MobileSAM-ViT-T $_{1024}$~\cite{zhang2023faster} & SA-1B & 8 &
    81.98 & 8.70 & 13.83 & 89.18 & 5.83 & 12.74 \\    
    EfficientSAM-ViT-T $_{1024}$~\cite{xiong2023efficientsam} & ImageNet+SA-1B & 6 &
    77.90 & 10.11 & 14.60 & 85.26 & 7.37 & 14.28 \\ 
    EfficientSAM-ViT-T $_{2048}$~\cite{xiong2023efficientsam} & ImageNet+SA-1B & 32 &
    74.20 & 9.47 & 13.13 & 84.10 & 8.00 & 14.37 \\ 
    EfficientSAM-ViT-S $_{1024}$~\cite{xiong2023efficientsam} & ImageNet+SA-1B & 7 &
    79.01 & 8.84 & 13.18 & 87.55 & 6.37 & 12.26 \\ 
    EfficientSAM-ViT-S $_{2048}$~\cite{xiong2023efficientsam} & ImageNet+SA-1B & 60 &
    74.91 & 8.27 & 11.97 & 85.17 & 6.86 & 12.49 \\ 
    SegNext (SA$\times$1) ViT-B $_{1024}$~\cite{liu2024rethinking} & COCO+LVIS & 42 &
    85.41 & 7.47 & 11.94 & 90.13 & 5.46 & 13.31 \\    
    SegNext (SA$\times$2) ViT-B $_{1024}$~\cite{liu2024rethinking} & COCO+LVIS & 58 &
    85.71 & 7.18 & 11.52 & 89.85 & 5.34 & 12.80 \\
    \midrule
    % \textit{Trained on HQSeg44K (HQ)} \\
    HQ-SAM-ViT-B $_{1024}$~\cite{ma2023segment} & SA-1B+HQ & 9 &
    89.85 & 6.49 & 10.79 & 91.77 & 5.26 & 10.00 \\
    SegNext (SA$\times$2) ViT-B $_{1024}$~\cite{liu2024rethinking} & COCO+LVIS+HQ & 58 &
    91.75 & 5.32 & 9.42 & 91.87 & 4.43 & 10.73 \\
    HRSAM-ViT-B $_{1024}$ (Ours) & COCO+LVIS+HQ 
    & 10 & 90.84 & 5.90 & 9.95 & 90.94 & 5.10 & 11.98 \\
    HRSAM-ViT-B $_{2048}$ (Ours) & COCO+LVIS+HQ 
    & 17 & 91.93 & 5.22 & 8.81 & 91.78 & 4.75 & 9.67 \\
    HRSAM$^{++}$-ViT-B $_{1024}$ (Ours) & COCO+LVIS+HQ 
    & 12 & 90.94 & 5.87 & 9.89 & 91.06 & 4.92 & 11.93 \\
    HRSAM$^{++}$-ViT-B $_{2048}$ (Ours) & COCO+LVIS+HQ 
    & 21 & 92.61 & 4.87 & 8.38 & 92.11 & 4.57 & 9.59 \\
    \midrule
    SAM-ViT-H $_{1024}$ (Teacher)~\cite{kirillov2023segany} & SA-1B & 29 &
    87.21 & 6.85 & 11.57 & 90.82 & 5.20 & 10.04 \\   
    HRSAM-ViT-B $_{1024}$ (Ours) & $^\dagger$COCO+LVIS {\small$_{\textbf{w/o labs}}$} & 10 &
    85.86 & 7.66 & 12.54 & 89.54 & 5.48 & 11.61 \\   
    HRSAM-ViT-B $_{2048}$ (Ours) & $^\dagger$COCO+LVIS {\small$_{\textbf{w/o labs}}$} & 17 &
    87.31 & 6.76 & 11.17 & 90.36 & 5.39 & 10.15 \\  
    HRSAM$^{++}$-ViT-B $_{1024}$ (Ours) & $^\dagger$COCO+LVIS {\small$_{\textbf{w/o labs}}$} & 12 &
    86.25 & 7.47 & 12.33 & 90.06 & 5.41 & 11.35 \\   
    HRSAM$^{++}$-ViT-B $_{2048}$ (Ours) & $^\dagger$COCO+LVIS {\small$_{\textbf{w/o labs}}$} & 21 &
    88.66 & 6.11 & 10.48 & 90.78 & 5.08 & 9.73 \\ 
    \midrule
    HRSAM-ViT-B $_{1024}$ (Ours) & $^\dagger$COCO+LVIS {\small$_{\textbf{w/o labs}}$}+HQ & 10 & 91.81 & 5.42 & 9.27 & 91.34 & 4.82 & 11.86 \\
    HRSAM-ViT-B $_{2048}$ (Ours) & $^\dagger$COCO+LVIS {\small$_{\textbf{w/o labs}}$}+HQ & 17 & \textbf{93.34} & 4.37 & 7.86 & 92.63 & 4.22 & 8.83 \\
    HRSAM$^{++}$-ViT-B $_{1024}$ (Ours) & $^\dagger$COCO+LVIS {\small$_{\textbf{w/o labs}}$}+HQ & 12 &
    91.84 & 5.32 & 9.18 & 91.25 & 5.02 & 11.64 \\
    HRSAM$^{++}$-ViT-B $_{2048}$ (Ours) & $^\dagger$COCO+LVIS {\small$_{\textbf{w/o labs}}$}+HQ & 21 &
    93.32 & \textbf{4.20} & \textbf{7.79} & \textbf{92.73} & \textbf{4.12} & \textbf{8.72} \\
    \bottomrule
  \end{tabu}}
  \vspace{-0.7em}
  \caption{Quantitative results. $\dagger$ indicates models trained by SAM-distillation without labels. Scaling to $2048^2$ allows HRSAM and HRSAM++ to outperform the previous SOTA SegNext in both performance and inference time under fair training on COCO+LVIS+HQ. SAM-distilled HRSAM models scaled to $2048^2$ also surpass their teacher model SAM-ViT-Huge in speed and performance. Further HQ finetuning of HRSAM++ achieves top segmentation performance.}
  \label{tab:quantitative_comparison}
  \vspace{-0.8em}
\end{table*}

\begin{table*}[!t]
  \centering
  \resizebox{\textwidth}{!}{
  \begin{tabu}[c]{l c c c c c c c c}
    \toprule
    \multirow{2}{*}{\textbf{Model}} &
    \multicolumn{2}{c}{\small\textbf{Img Size $\in$ (0, 1024]}} &
    \multicolumn{2}{c}{\small\textbf{Img Size $\in$ (1024, 2048]}} &
    \multicolumn{2}{c}{\small\textbf{Img Size $\in$ (2048, 3072]}} &
    \multicolumn{2}{c}{\small\textbf{Img Size $\in$ (3072, 4096]}} \\
    \cmidrule(lr){2-3} 
    \cmidrule(lr){4-5} 
    \cmidrule(lr){6-7} 
    \cmidrule(lr){8-9}
    & 
    \textbf{NoC90} $\downarrow$ & \textbf{NoC95} $\downarrow$ & \textbf{NoC90} $\downarrow$ & \textbf{NoC95} $\downarrow$ & \textbf{NoC90} $\downarrow$ & \textbf{NoC95} $\downarrow$ & \textbf{NoC90} $\downarrow$ & \textbf{NoC95} $\downarrow$ \\
    \midrule
    SAM-ViT-H $_{1024}$~\cite{kirillov2023segany}
    & 5.23 & 10.27 & 7.94 & 13.00 & 7.67 & 11.87 & 7.79 & 12.45 \\
    \midrule
    HRSAM-ViT-B $_{1024}$
    & 5.80 & 10.99 & 8.60 & 13.82 & 8.19 & 12.69 & 9.47 & 13.25 \\
    HRSAM-ViT-B $_{2048}$
    & \textbf{5.67} & \textbf{10.16} & \textbf{7.87} & \textbf{12.82} & \textbf{7.08} & \textbf{11.00} & \textbf{7.28} & \textbf{11.56} \\
    HRSAM-ViT-B $_{3072}$
    & 6.91 & 11.24 & 8.76 & 13.96 & 7.75 & 11.59 & 7.76 & 12.24 \\
    HRSAM-ViT-B $_{4096}$
    & 8.54 & 13.25 & 10.69 & 15.69 & 8.99 & 13.20 & 8.76 & 13.56 \\
    \midrule
    HRSAM$^{++}$-ViT-B $_{1024}$
    & 5.58 & 10.78 & 8.53 & 13.65 & 7.97 & 12.30 & 9.18 & 13.19 \\
    HRSAM$^{++}$-ViT-B $_{2048}$
    & \textbf{4.95} & \textbf{9.42} & \textbf{7.18} & \textbf{12.15} & 6.46 & 10.33 & 6.65 & 11.14 \\
    HRSAM$^{++}$-ViT-B $_{3072}$
    & 5.10 & 9.87 & 7.28 & 12.41 & \textbf{6.27} & \textbf{9.98} & \textbf{6.49} & \textbf{10.74} \\
    HRSAM$^{++}$-ViT-B $_{4096}$
    & 5.57 & 10.82 & 8.28 & 13.20 & 6.36 & 10.68 & 6.79 & 11.33 \\
    \bottomrule
  \end{tabu}}
  \vspace{-0.7em}
  \caption{Input size scalability analysis. The metrics NoC@90 and NoC@95 are measured over the HQSeg44K dataset.}
  \vspace{-1em}
  \label{tab:multiscale}
\end{table*}

\subsection{Experimental Setting}
\label{subsec:experimentalsetting}

We provide the model and training/testing details in Appendix~\ref{app:experimentaldetails}.

\noindent\textbf{Datasets.} 
We use COCO~\cite{TsungYiLin2014MicrosoftCC}, LVIS~\cite{AgrimGupta2019LVISAD} and HQSeg-44K~\cite{sam_hq} for training. Evaluation is conducted exclusively on HQSeg44K’s validation set and DAVIS~\cite{FedericoPerazzi2016ABD}, aligning with high-precision interactive segmentation benchmarks~\cite{liu2024rethinking}.

\noindent\textbf{Training strategy.}
We adopt three training settings. 
First, following the prior method~\cite{liu2024rethinking}, we train on COCO~\cite{TsungYiLin2014MicrosoftCC} and LVIS~\cite{AgrimGupta2019LVISAD} with click simulation and normalized focal loss (NFL), then finetune on HQSeg44K~\cite{sam_hq} for fair comparison. Second, we distill from SAM-ViT-Huge~\cite{sam_hq} on COCO and LVIS without labels, using a simple MSE loss for direct comparison with SAM. Finally, we finetune the distilled weights on HQSeg44K with simulated clicks and NFL.
Details are provided in Appendix~\ref{app:experimentaldetails}.

% \vspace{-0.5em}
\noindent\textbf{Evaluation.}
We evaluate HRSAMs under the standard protocol~\cite{XiChen2022FocalClickTP,QinLiu2022SimpleClickII,huang2023interformer}. Performance is measured by 5-mIoU (mean IoU after five clicks) and NoC@90/95, \ie the average clicks to reach $90\%/95\%$ IoU. Speed is measured as Seconds Per Click (20-SPC) on GPUs, \ie the average inference time per click over 20 clicks. Appendix~\ref{app:experimentaldetails} provides details.

\begin{table*}[!t]
  \centering
  % \vspace{-0.525em}
  \resizebox{\textwidth}{!}{
  \begin{tabu}[c]{c c c c c c c c c c c}
    \toprule
    \multirow{2}{*}{\textbf{Global}} & 
    \textbf{Flash} &
    \textbf{KV-only} &
    \textbf{Cycle} &
    \textbf{Multi} &
    \multicolumn{3}{c}{\textbf{1024$\times$1024 Image}} & 
    \multicolumn{3}{c}{\textbf{2048$\times$2048 Image}} \\
    \cmidrule(lr){6-8} \cmidrule(lr){9-11}
    & \textbf{Swin} & \textbf{Padding} & \textbf{Scan} & \textbf{Scale} & 
    Lat (ms) $\downarrow$ & NoC90 $\downarrow$ & NoC95 $\downarrow$ &
    Lat (ms) $\downarrow$ & NoC90 $\downarrow$ & NoC95 $\downarrow$ \\
    \midrule
    \ding{51} & & & & &  
    104 & \textbf{7.47} & \textbf{12.30} & 1338 & 6.89 & 11.61 \\
    & \ding{51} & & & & 
    37 & 7.68 & 12.71 & 122 & 6.82 & 11.44 \\
    & \ding{51} & \ding{51} & & & 
    39 & 7.68 & 12.69 & 130 & 6.81 & 11.36 \\
    & \ding{51} & & \ding{51} & & 
    52 & 7.59 & 12.59 & 178 & 6.87 & 11.34 \\
    & \ding{51} & \ding{51} & \ding{51} & & 
    53 & 7.66 & 12.54 & 185 & 6.76 & 11.17 \\
    & \ding{51} & \ding{51} & \ding{51} & \ding{51} &
    90 & \textbf{7.47} & 12.33 & 269 & \textbf{6.11} & \textbf{10.48} \\
    \bottomrule
\end{tabu}}
  \vspace{-0.5em}
  \caption{Ablation study of proposed modules. Latency is measured on the image encoder, and performance on HQSeg44K.}
  \label{tab:ablation}
  \vspace{-0.4em}
\end{table*}

\begin{figure*}[!ht]
    \centering
    \vspace{-0.2em}
    \includegraphics[width=0.96\textwidth]{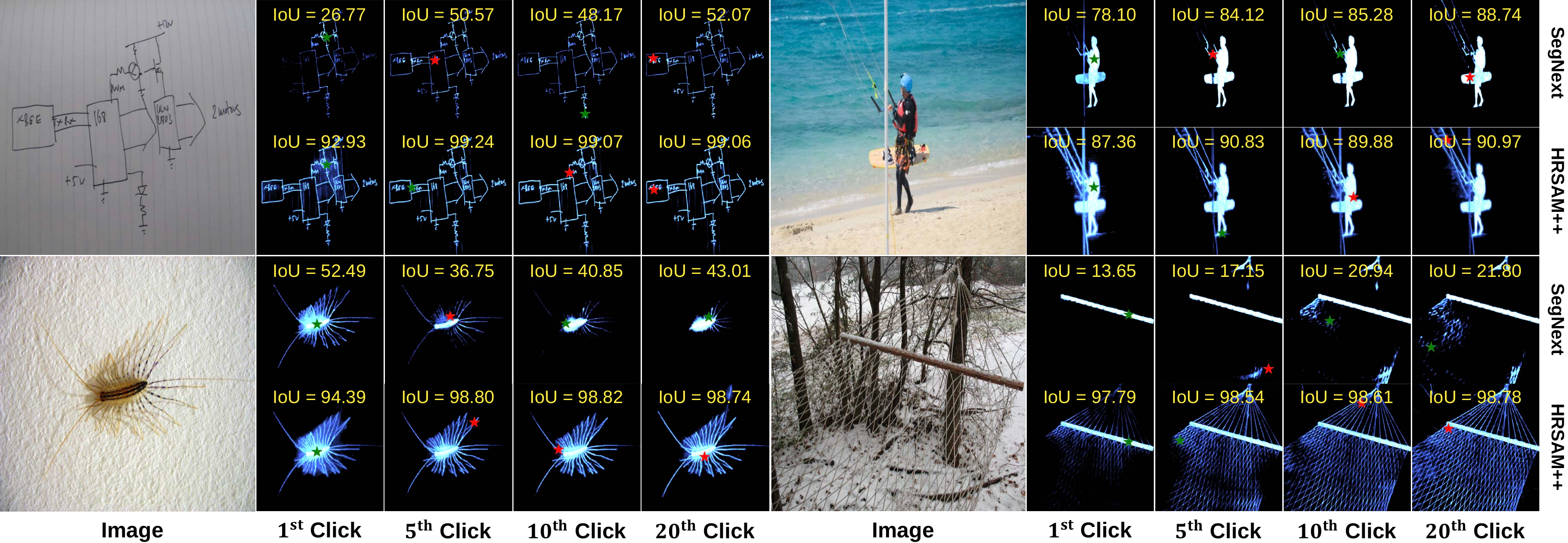}
    \vspace{-0.5em}
    \caption{Qualitative comparison of our HRSAM++ and previous SOTA SegNext. 
    }
    \label{fig:qualitativeresults}
    \vspace{-1.3em}
\end{figure*}

\subsection{Main Results}
\label{subsec:mainresult}

As shown in Table~\ref{tab:quantitative_comparison}, 
under fair training conditions, HRSAM and HRSAM++ surpass the previous SOTA SegNext~\cite{liu2024rethinking} with substantially lower 20-SPC (inference time). 
When trained via SAM-distillation on $1024^2$ inputs, HRSAM and HRSAM++ outperform their teacher model SAM-ViT-Huge at $2048^2$ testing input, retaining a notable speed advantage. Building on this, the HQ-finetuned HRSAM++ achieves significantly higher performance than  the previous SOTA~\cite{liu2024rethinking} on high-precision datasets HQSeg44K and DAVIS.

\subsection{Scale Analysis}
\label{subsec:scaleanlysis}
 
To analyze the scalability of HRSAM with larger input sizes, we evaluate HRSAM and HRSAM++ at $1024^2$, $2048^2$, $3072^2$ and $4096^2$ resolutions on the HQSeg44K dataset, focusing on NoC@90/95.
As shown in Table~\ref{tab:multiscale}, 
both HRSAM and HRSAM++ scale effectively from a $1024^2$ training resolution to $2048^2$ testing size. However, for larger inputs like $3072^2$ and $4096^2$, HRSAM's performance declines, likely due to insufficient resizing precision and the limits of length extrapolation. In contrast, HRSAM++ maintains improved performance for images resized to $3072^2$, demonstrating the enhanced visual length extrapolation of the anchor map.

\subsection{Ablation Study}
\label{sec:ablationstudy}

In the ablation study, we train HRSAM and its variants using the same distillation protocol as in the main experiments to evaluate the effects of the proposed modules.
As shown in Table~\ref{tab:ablation}, Global attention achieves strong performance with $1024^2$ inputs due to its high capacity but exhibits an imbalanced trade-off between performance and latency at $2048^2$ inputs. In contrast, the HRSAM variants show a better balance at larger scales. Notably, HRSAM++ with Multi-scale anchor map achieves significant performance gains at $2048^2$, underscoring its effectiveness in enhancing scalability for high-resolution inputs.

\subsection{Qualitative Results}
\label{subsec:qualitativeresults}

Figure~\ref{fig:qualitativeresults} presents a qualitative comparison between our HRSAM++ and the previous SOTA SegNext~\cite{liu2024rethinking} over challenging samples with thin edges, clearly demonstrating the superior performance of HRSAM++ in high-precision segmentation tasks.
More results are provided in Appendix~\ref{app:results}.

%% file: sec/5_conclusion.tex
\section{Conclusion}
In this paper, we propose HRSAM to address SAM's limitations in handling high-resolution images for high-precision interactive segmentation. Built on the proposed FLA framework, HRSAM integrates Flash Swin attention, a KV-only padding mechanism and the Cycle-scan module to enhance computational efficiency and receptive field. The further proposed HRSAM++ incorporates an anchor map, providing multi-scale data augmentation and a larger receptive field. Our HRSAM models establish new benchmarks in high-precision interactive segmentation. 

%% file: sec/X_suppl.tex
\clearpage
\appendix
\setcounter{page}{1}
\maketitlesupplementary

\section{Shift Window Index Mapping}
\label{app:shiftwindow}
To specify the index mapping for the shift window within the proposed FLA framework, we previously define the extended index mappings in the main text as follows:
\begingroup
\setlength{\abovedisplayskip}{0.5em}
\setlength{\belowdisplayskip}{0.5em}
\begin{equation}
\scalebox{0.73}{$
\begin{split}
    &\mathcal{I}_{\text{val}}^\prime(j; h, w) = \left(\left\lfloor \frac{j - 1}{w + k}\right\rfloor + 1 - \frac{k}{2}, (j - 1 \bmod (w + k)) + 1 - \frac{k}{2}\right),\\
    &\mathcal{I}_{\text{win}}^\prime(y, x; h, w, k) =  \mathcal{I}_{\text{win}}\left(y + \frac{k}{2}, x + \frac{k}{2}; h + k, w + k, k\right).
\end{split}
$}
\end{equation}
\endgroup
Then, we use the composition of the inverse mappings to derive the final index mapping, \ie $\mathcal{I}_{\text{swin}}^\prime = \mathcal{I}_{\text{val}}^{\prime-1} \circ \mathcal{I}_{\text{win}}^{\prime-1}$, which maps from $I^\prime = \{1, 2, \ldots, (h + k)(w + k)\}$ onto itself. A subset of these indices corresponds to virtual tokens, which are removed to obtain a mapping that represents only the original indices. We identify these indices by $\mathcal{I}_{\text{win}}^\prime$, \ie
\begingroup
\setlength{\abovedisplayskip}{0.5em}
\setlength{\belowdisplayskip}{0.5em}
\begin{equation}
\scalebox{0.85}{$
\begin{split}
    I_v = \mathcal{I}_{\text{win}}^{\prime}\left(\left\{(y, x) \,|\, y \notin \{1, 2, \ldots, h\} \text{ or } x \notin \{1, 2, \ldots, w\}\right\}\right).
\end{split}
$}
\end{equation}
\endgroup
The subsets $I^\prime \setminus I_v, I^\prime \setminus\mathcal{I}_{\text{swin}}^\prime(I_v)$ correspond to the set of non-virtual token indices within the domain/range of $\mathcal{I}_{\text{swin}}^\prime$. We further define the shift index mapping to transform these indices back onto the original domain $\{1, 2, \ldots, L\}$ as follows:
\begingroup
\setlength{\abovedisplayskip}{0.5em}
\setlength{\belowdisplayskip}{0.5em}
\begin{equation}
\scalebox{0.95}{$
\begin{split}
    &\mathcal{S}_A: I^\prime \setminus A \rightarrow \{1, 2, \ldots, L\}, \\
    &\mathcal{S}_{A}(i) < \mathcal{S}_{A}(j) \iff i < j, \forall i, j \in I^\prime \setminus A,
\end{split}
$}
\end{equation}
\endgroup
where $A$ is any set of indices satisfying $|I^\prime \setminus A| = L$. Therefore, the final shift window index mapping is defined as the composition
\begingroup
\setlength{\abovedisplayskip}{0.5em}
\setlength{\belowdisplayskip}{0.5em}
\begin{equation}
\scalebox{0.95}{$
\begin{split}
    \mathcal{I}_{\text{swin}} = \mathcal{S}_{\mathcal{I}_{\text{swin}}^\prime(I_v)}
    \circ
    \mathcal{I}_{\text{swin}}^\prime
    \circ
    \mathcal{S}_{I_v}^{-1}.
\end{split}
$}
\label{eq:finalswin}
\end{equation}
\endgroup
Then, the subset partition length $L_i$ is defined as
\begingroup
\setlength{\abovedisplayskip}{0.5em}
\setlength{\belowdisplayskip}{0.5em}
\begin{equation} 
\begin{split} 
    L_i = \left| \mathcal{I}_{\text{swin}}^\prime \left(\{ 1, 2, \ldots, ik^2\}\right) \setminus \mathcal{I}_{\text{swin}}^\prime(I_v) \right|, 
\end{split} 
\label{eq:finalswinlengths}
\end{equation}
\endgroup
for $1 \le i \le (h + k)(w + k) / k^2$.
Eq~\ref{eq:finalswin} and Eq~\ref{eq:finalswinlengths} can be simplified into the code implementation shown in Listing~\ref{alg:shiftwindow}.

\begin{listing}[!ht]
\caption{Pseudo code for shift window index mapping based on PyTorch. }
\label{alg:shiftwindow}
\begin{minted}{python}
def make_index_mapping_of_shift_window(h, w, k, INF):
    """
    :param h: int, height of the image embedding.
    :param w: int, width of the image embedding.
    :param k: int, window size; shift = k // 2.
    :param INF: a large number used to
                mark virtual tokens.
    :return:
        - indices: shift window index with shape
                   (h * w,).
                   For x of shape (h * w, C),
                   x[indices] rearranges the
                   token order.
        - lengths: subset lengths
                   within each window.
        - inv_indices: inverse index for
                       restoring the original order,
                       i.e.,
                       x == x[indices][inv_indices].
    """
    # left padding of k // 2
    # right padding of k // 2 tokens
    hs = torch.arange(0 - k // 2, h + k // 2)
    # top padding of k // 2
    # bottom padding of k // 2 tokens
    ws = torch.arange(0 - k // 2, w + k // 2)

    # use a large negative number to
    # mark virtual tokens
    hs[:k // 2] = -INF
    hs[-k // 2:] = -INF
    ws[:k // 2] = -INF
    ws[-k // 2:] = -INF

    # make 2D coordinates
    ys, xs = torch.meshgrid(hs, ws)
    # transform to 1D coordinates
    coords = ys * w + xs

    # make indices within the same window
    # contiguously arranged
    coords = coords.reshape((h + k) // k, k,
                            (w + k) // k, k)
    coords = coords.permute(0, 2, 1, 3)
    coords = coords.reshape(-1, k * k)

    # count each window's valid tokens
    lengths = (coords >= 0).sum(dim=-1).tolist()

    # remove the virtual tokens' indices
    indices = coords.reshape(-1)
    indices = indices[indices >= 0]
    return indices, lengths, indices.argsort()
\end{minted}
\vspace{-1em}
\end{listing}
\vspace{-1em}

\section{KV-only Padding Index Mapping}
\label{app:kvpad}
To accommodate a learnable token appended to the end of the image token sequence, we extend the original tokens $\mathbf{X} \in \mathbb{R}^{L \times C}$ to $\mathbf{X}^\prime \in \mathbb{R}^{(L + 1) \times C}$. Based on this, we construct the transitional mapping
$\mathcal{T}_{\text{pad}}: I^\prime \rightarrow \{1, 2, \ldots, L, L + 1\}$, defined as
\begingroup
\setlength{\abovedisplayskip}{0.5em}
\setlength{\belowdisplayskip}{0.5em}
\begin{equation}
\mathcal{T}_{\text{pad}}(j) = \left\{
\begin{split}
    \mathcal{S}_{\mathcal{I}_{\text{swin}}^\prime(I_v)}(j), \quad&j \notin \mathcal{I}_{\text{swin}}^\prime(I_v), \\
    L + 1, \quad&j \in \mathcal{I}_{\text{swin}}^\prime(I_v). \\
\end{split}
\right.
\end{equation}
\endgroup
This transition mapping derives the KV-only padding index mapping as follows:
\begingroup
\setlength{\abovedisplayskip}{0.5em}
\setlength{\belowdisplayskip}{0.5em}
\begin{equation}
\begin{split}
    \mathcal{I}_{\text{pad}} = \mathcal{T}_{\text{pad}}\circ\mathcal{I}_{\text{swin}}^\prime.
\end{split}
\end{equation}
\endgroup
The subset partition length is thus defined as:
\begingroup
\setlength{\abovedisplayskip}{0.5em}
\setlength{\belowdisplayskip}{0.5em}
\begin{equation}
\begin{split}
    L_i = ik^2, \quad 1 \le i \le (h + k)(w + k) / k^2.
\end{split}
\end{equation}
\endgroup
Similarly, this index mapping can be simplified into the code implementation shown in Listing~\ref{alg:kvpad}.

\begin{listing}[!ht]
\caption{Pseudo code for KV-padding index mapping based on PyTorch.}
\label{alg:kvpad}
\begin{minted}{python}
def make_index_mapping_of_kv_padding(h, w, k, INF):
    """
    :param h: int, height of the image embedding.
    :param w: int, width of the image embedding.
    :param k: int, window size; shift = k // 2.
    :param INF: a large number used to
                mark virtual tokens.
    :return:
        - indices: shift window index with shape
                   ((h + k) * (w + k),),
                   used to rearrange tokens for
                   x of shape (h * w + 1, C)
        - lengths: subset lengths
                   within each window.
    """
    # left padding of k // 2
    # right padding of k // 2 tokens
    hs = torch.arange(0 - k // 2, h + k // 2)
    # top padding of k // 2
    # bottom padding of k // 2 tokens
    ws = torch.arange(0 - k // 2, w + k // 2)

    # use a large negative number to
    # mark virtual tokens
    hs[:k // 2] = -INF
    hs[-k // 2:] = -INF
    ws[:k // 2] = -INF
    ws[-k // 2:] = -INF

    # make 2D coordinates
    ys, xs = torch.meshgrid(hs, ws)
    # transform to 1D coordinates
    coords = ys * w + xs

    # make indices within the same window
    # contiguously arranged
    coords = coords.reshape((h + k) // k, k,
                            (w + k) // k, k)
    coords = coords.permute(0, 2, 1, 3)
    indices = coords.reshape(-1)

    # count each window's tokens
    lengths = ((h + k) * (w + k)) // k ** 2 * \
              [k ** 2]

    # set the index of virtual tokens
    # to the padding token's index
    indices[indices < 0] = h * w
    return indices, lengths
\end{minted}
\vspace{-1em}
\end{listing}

\section{Multi-scale Index Mapping}
\label{app:multiscale}

\subsection{Non-padding index mapping}
For the multi-scale input $\mathbf{X} \in \mathbb{R}^{L \times C}$ with $L \!=\! HW / 16^2 + H_sW_s / 16^2$, we define the index mapping under the FLA to achieve multi-scale attention, through the shift mapping $\mathcal{I}_s(j) = j - HW / 16^2$:
\begingroup
\setlength{\abovedisplayskip}{0.5em}
\setlength{\belowdisplayskip}{0.5em}
\begin{equation}
\scalebox{0.9}{$
\mathcal{I}_{\text{ms}}(j) = 
\left\{
\begin{split} 
&\mathcal{I}_{\text{swin}}\left(j; \frac{H}{16}, \frac{W}{16}\right)
, j \leq \frac{HW}{16^2}, \\
&\mathcal{I}_s^{-1}\left(\mathcal{I}_{\text{swin}}\left(\mathcal{I}_s(j); \frac{H_s}{16}, \frac{W_s}{16}\right)\right)
, j > \frac{HW}{16^2}.
\end{split}\right.
$}
\end{equation}
\endgroup
The mapping can be simplified into the code of Listing~\ref{alg:multiscale}.

\begin{listing}[!ht]
\caption{Pseudo code for multi-scale index mapping based on PyTorch.}
\label{alg:multiscale}
\begin{minted}{python}
def make_index_mapping_of_multi_scale(h, w,
                                      hs, ws,
                                      k, INF):
    """
    :param h: int, height of the
              original image embedding.
    :param w: int, width of the
              original image embedding.
    :param hs: int, height of the
               downsampled image embedding.
    :param ws: int, width of the
               downsampled image embedding.
    :param k: int, window size; shift = k // 2.
    :param INF: a large number used to
                mark virtual tokens.
    :return:
        - indices: shift window index with shape
                   (h * w + hs * ws, ).
                   For x of shape
                   (h * w + hs * ws, C),
                   x[indices] rearranges the
                   token order.
        - lengths: subset lengths
                   within each window.
        - inv_indices: inverse index for
                       restoring the original order,
                       i.e.,
                       x == x[indices][inv_indices].
    """
    # get the indices of the original embeddings
    ori_indices, ori_lengths, _ = \
        make_index_mapping_of_shift_window(
            h, w, k, INF)
    # get the indices of the downsampled embeddings
    ds_indices, ds_lengths, _ = \
        make_index_mapping_of_shift_window(
            hs, ws, k, INF)

    # concatenate the indices
    indices = torch.cat([ori_indices,
                         ds_indices + h * w])
    lengths = ori_lengths + ds_lengths
    return indices, lengths, indices.argsort()
\end{minted}
\vspace{-1em}
\end{listing}

\subsection{KV-only padding index mapping}

For the multi-scale KV-only padding mechanism, a shared learnable token is appended as the padding token at the end of the entire input $\mathbf{X}$. This ensures consistent alignment across scales, where both the main scale and the anchor map's scale mappings are adjusted to point to this shared padding token. The implementation is shown in Listing~\ref{alg:multiscalekvpad}.

\begin{listing}[!ht] 
\caption{Pseudo code for multi-scale KV-only padding with a shared learnable token based on PyTorch.}
\label{alg:multiscalekvpad} 
\begin{minted}{python}
def make_index_mapping_of_ms_kv_padding(h, w,
                                        hs, ws,
                                        k, INF):
    """
    :param h: int, height of the
              original image embedding.
    :param w: int, width of the
              original image embedding.
    :param hs: int, height of the
               downsampled image embedding.
    :param ws: int, width of the
               downsampled image embedding.
    :param k: int, window size; shift = k // 2.
    :param INF: a large number used to
                mark padding tokens.
    :return:
        - indices: shift window index with shape
                   ((h + k) * (w + k) +
                    (hs + k) * (ws + k), ).
                   For x of shape
                   (h * w + hs * ws + 1, C),
                   x[indices] rearranges the
                   token order.
        - lengths: subset lengths
                   within each window.
    """
    # get the indices of the original embeddings
    ori_indices, ori_lengths = \
        make_index_mapping_of_kv_padding(
            h, w, k, INF)
    # get the indices of the downsampled embeddings
    ds_indices, ds_lengths = \
        make_index_mapping_of_kv_padding(
            hs, ws, k, INF)

    # reset the index of padding tokens
    last = h * w + hs * ws
    ori_indices[ori_indices == h * w] = last

    # concatenate the indices
    indices = torch.cat([ori_indices,
                         ds_indices + h * w])
    lengths = ori_lengths + ds_lengths
    return indices, lengths
\end{minted}
\vspace{-1em} 
\end{listing}

\section{Experimental Details}
\label{app:experimentaldetails}

\subsection{Datasets} 
We adopt COCO~\cite{TsungYiLin2014MicrosoftCC}, LVIS~\cite{AgrimGupta2019LVISAD} and HQSeg44K~\cite{sam_hq} datasets for our experiments. The COCO dataset includes 118K training images (1.2M instances), while LVIS shares the same images as COCO but provides higher-quality segmentation annotations. HQSeg44K contains approximately 44K training images and over 1,500 validation images. We train our models on the combined training sets of COCO, LVIS and HQSeg44K and evaluate them on the validation set of HQSeg44K and 345 images from the DAVIS dataset~\cite{FedericoPerazzi2016ABD}, following the protocol of previous high-precision interactive segmentation studies~\cite{liu2024rethinking}.

\subsection{Model details}
We implement the image encoder of HRSAM based on ViT-Base~\cite{AlexeyDosovitskiy2020AnII}. The input image is patchified into image tokens through a $16\times16$ patch embedding process. HRSAM-ViT-Base consists of 12 blocks evenly distributed across 4 stages. The Cycle-scan module is added before the last block of each stage. Each block processes image tokens with a dimension of 768. The inputs and outputs of each stage have the same dimension of 768. The outputs of all stages are summed and passed through a convolution layer to reduce the dimension to 256. This final output is compatible with the SAM decoder.
Each block uses Flash Swin attention and a standard FFN module. The attention mechanism has 12 heads, and the FFN has a hidden layer dimension of $768 \times 4$. The Cycle-scan module uses a Mamba SSM~\cite{gu2023mamba} with a hidden dimension of 32.
The weights of HRSAM-ViT-Base are initialized using MAE-pretrained weights~\cite{KaimingHe2021MaskedAA}, following previous works~\cite{QinLiu2022SimpleClickII,huang2023interformer,liu2024rethinking}.

For HRSAM++, the input image is downsampled to an additional $512 \times 512$ resolution as the anchor map. The anchor map passes through the same 4 stages as the original scale, and the weights of these stages are shared across scales. Before the output of each stage, the image tokens from both scales are concatenated and processed by an additional Cycle-scan module configured identically to the Cycle-scan module in HRSAM. This module performs multi-scale feature fusion. After fusion, the tokens are split back into their respective scales. The output from the anchor map is upsampled to align with the original scale. The results from both scales are concatenated along the channel dimension, producing an output dimension of $768 \times 2$. The outputs from the 4 stages are summed and passed through a convolution layer to reduce the dimension to 256, consistent with the original HRSAM. 

Further implementation details are presented in the \href{https://github.com/YouHuang67/High-Resolution-Segment-Anything.git}{code}.

\begin{table*}[!ht]
  \centering
  \resizebox{\textwidth}{!}{
  \begin{tabu}[c]{l c c c c c c c}
    \toprule
    \multirow{2}{*}{\textbf{Model}} &
    \textbf{Inference Time} &
    \multicolumn{3}{c}{\textbf{HQSeg44K} {\small$_{\textbf{Max H/W} > 4000}$}} & 
    \multicolumn{3}{c}{\textbf{DAVIS} {\small$_{\textbf{Max H/W} < 1000}$}} \\
    \cmidrule(lr){3-5} \cmidrule(lr){6-8}
    & \textbf{20-SPC (ms) $\downarrow$} & 
    5-mIoU $\uparrow$ & NoC90 $\downarrow$ & NoC95 $\downarrow$ &
    5-mIoU $\uparrow$ & NoC90 $\downarrow$ & NoC95 $\downarrow$ \\
    \midrule
    SAM-ViT-H $_{1024}$ (Teacher)~\cite{kirillov2023segany} & 29 &
    87.21 & 6.85 & 11.57 & 90.82 & 5.20 & 10.04 \\   
    \midrule
    HRSAM-ViT-B $_{1024}$ & 10 &
    85.86 & 7.66 & 12.54 & 89.54 & 5.48 & 11.61 \\ 
    HRSAM-ViT-B $_{1024\uparrow2048}$ & 17 &
    87.18 & 6.82 & 11.31 & 90.38 & 5.40 & 10.20 \\  
    HRSAM-ViT-B $_{2048}$ & 17 &
    87.31 & 6.76 & 11.17 & 90.36 & 5.39 & 10.15 \\  
    HRSAM$^{++}$-ViT-B $_{1024}$ & 12 &
    86.25 & 7.47 & 12.33 & 90.06 & 5.41 & 11.35 \\  
    HRSAM$^{++}$-ViT-B $_{1024\uparrow2048}$ & 21 &
    88.54 & 6.24 & 10.69 & 90.48 & 5.14 & 9.86 \\  
    HRSAM$^{++}$-ViT-B $_{2048}$ & 21 &
    88.66 & 6.11 & 10.48 & 90.78 & 5.08 & 9.73 \\ 
    \bottomrule
  \end{tabu}
  }
  \caption{Upsampling Evaluation Results. Performance of HRSAMs is evaluated under two input handling modes, resizing and padding to $1024^2$ or $2048^2$ resolutions, and further upsampling $1024^2$ inputs to $2048^2$ ($1024 \uparrow 2048$). Upsampling improves performance compared to $1024^2$ inputs but performs slightly worse than $2048^2$ resized inputs.}
  \label{tab:addscaleresults}
\end{table*}

\subsection{Training strategy}

In the training process, we adopt three settings to train HRSAMs. First, under fair conditions, we train from scratch using the same protocol as previous works on interactive segmentation~\cite{liu2024rethinking}. This involves click simulation and normalized focal loss (NFL)~\cite{KonstantinSofiiuk2021RevivingIT,huang2023interformer,liu2024rethinking}. The training starts with 160k iterations on the COCO and LVIS datasets, followed by finetuning on the HQSeg44K dataset until convergence, consistent with prior studies~\cite{liu2024rethinking}.
Second, we adopt SAM-distillation to train HRSAMs using image embeddings from the teacher model SAM-ViT-Huge. Specifically, SAM-ViT-Huge is used to generate image embeddings on COCO and LVIS images, which are stored offline. During the training, these embeddings are used as the target. HRSAMs generate the embeddings on the same images, and the difference between HRSAM embeddings and the precomputed SAM embeddings is minimized using the simple MSE loss:
\begingroup
\setlength{\abovedisplayskip}{0.5em}
\setlength{\belowdisplayskip}{0.5em}
\begin{equation} 
\scalebox{0.82}{$
\min_{\mathbf{\theta}} \sum_i \left|\left| \text{SAM-ViT-Huge}(\mathbf{x}_i) - \text{HRSAM-ViT-Base}(\mathbf{x}_i; \mathbf{\theta}) \right|\right|_2^2. 
$}
\end{equation}
\endgroup
This training process involves 160k iterations on COCO and LVIS. 
Finally, in the third training setting, the SAM-distilled model is further finetuned on the HQSeg44K dataset with segmentation annotations, following the same approach as the first setting.

All training settings share most of the configurations. The optimizer is AdamW with weight decay of 0.05 and betas of $(0.9, 0.999)$. We use a learning rate scheduler with a linear warm-up starting from $1 \!\times\! 10^{-6}$ and a polynomial decay starting from $1 \!\times\! 10^{-4}$. Training is conducted on 4 NVIDIA RTX 3090 GPUs with a batch size of 1 per GPU.

\subsection{Evaluation}
We evaluate HRSAMs against various methods~\cite{KonstantinSofiiuk2021RevivingIT,XiChen2022FocalClickTP,QinLiu2022SimpleClickII,huang2023interformer,liu2024rethinking,kirillov2023segany,zhang2023faster,xiong2023efficientsam}. For non-SAM models, we use their default input sizes. For SAM series models, we use both $1024^2$ and $2048^2$ inputs if the model supports $2048^2$ inputs. In testing, click simulation places clicks at the centers of erroneously predicted regions, with the binary label of each click determined by the maximum distance to the boundaries of false negative and false positive regions, aligning with the previous methods~\cite{XiChen2022FocalClickTP,QinLiu2022SimpleClickII,huang2023interformer}. Segmentation performance is measured using 5-mIoU and NoC metrics, following the previous study~\cite{liu2024rethinking}. The 5-mIoU represents the average IoU after the fifth click. The NoC metric indicates the average minimum clicks required to reach a specified IoU. We focus on NoC@90 and NoC@95 within 20 clicks. 

\subsection{Inference time}
Speed is quantified as Seconds Per Click (SPC) on GPUs, measuring the average inference latency per click over 20 clicks. For traditional interactive segmentation models, SPC is calculated as the time taken from input to output after each click. For SAM-like models, where the image encoder is only used once per image, we use the 20-SPC metric to ensure a fair comparison. This metric averages the encoder’s inference time across 20 clicks during standard testing. The SPC for SAM-like models thus includes the preprocessing time by the image encoder and subsequent operations.

Besides, the speed analysis in sections outside the main results focuses on the standalone latency of various HRSAM ablation structures. This latency only accounts for the inference time of the image encoder, which differs from SPC.

\section{Evaluation on Upsampled Inputs}
\label{app:results}

To explore the impact of upsampling on model performance, we evaluate HRSAMs under two input handling modes. In the first mode, input images of any size are resized and padded to either $1024^2$ or $2048^2$ resolutions for direct input into the model. In the second mode, images resized to $1024^2$ are further upsampled using linear interpolation to generate $2048^2$ inputs ($1024 \uparrow 2048$). This setup investigates whether such upsampling can enhance interactive segmentation performance, even without access to fine details in high-resolution images.

The results in Table~\ref{tab:addscaleresults} demonstrate that upsampling from $1024^2$ to $2048^2$ improves model performance. However, compared to directly using resized $2048^2$ inputs, the upsampling process loses some fine details, leading to slightly lower segmentation accuracy. Despite this, upsampled inputs perform significantly better than directly resized $1024^2$ inputs, confirming that HRSAM possesses sufficient extrapolation capability to generalize effectively to larger images.

Besides, with such upsampled $2048^2$ inputs, HRSAM++ achieves significantly better performance and lower latency than the teacher model SAM-ViT-Huge. This demonstrates HRSAM++'s ability to effectively utilize interpolated inputs without relying on additional fine details. These findings suggest that the low-resolution upsampling approach has untapped potential to enhance model performance while reducing dependence on high-resolution details.

%% file: main.bbl
\begin{thebibliography}{109}
\providecommand{\natexlab}[1]{#1}
\providecommand{\url}[1]{\texttt{#1}}
\expandafter\ifx\csname urlstyle\endcsname\relax
  \providecommand{\doi}[1]{doi: #1}\else
  \providecommand{\doi}{doi: \begingroup \urlstyle{rm}\Url}\fi

\bibitem[Acuna et~al.(2018)Acuna, Ling, Kar, and Fidler]{DavidAcuna2018EfficientIA}
David Acuna, Huan Ling, Amlan Kar, and Sanja Fidler.
\newblock Efficient interactive annotation of segmentation datasets with polygon-rnn++.
\newblock In \emph{2018 {IEEE} Conference on Computer Vision and Pattern Recognition, {CVPR} 2018, Salt Lake City, UT, USA, June 18-22, 2018}, pages 859--868. {IEEE} Computer Society, 2018.

\bibitem[Beyer et~al.(2023)Beyer, Izmailov, Kolesnikov, Caron, Kornblith, Zhai, Minderer, Tschannen, Alabdulmohsin, and Pavetic]{beyer2023flexivit}
Lucas Beyer, Pavel Izmailov, Alexander Kolesnikov, Mathilde Caron, Simon Kornblith, Xiaohua Zhai, Matthias Minderer, Michael Tschannen, Ibrahim Alabdulmohsin, and Filip Pavetic.
\newblock Flexivit: One model for all patch sizes.
\newblock In \emph{CVPR}, 2023.

\bibitem[Cai et~al.(2020)Cai, Gan, Wang, Zhang, and Han]{8}
Han Cai, Chuang Gan, Tianzhe Wang, Zhekai Zhang, and Song Han.
\newblock Once for all: Train one network and specialize it for efficient deployment.
\newblock \emph{ICLR}, 2020.

\bibitem[Chen et~al.(2024{\natexlab{a}})Chen, Song, Han, Xia, and Yokoya]{chen2024changemamba}
Hongruixuan Chen, Jian Song, Chengxi Han, Junshi Xia, and Naoto Yokoya.
\newblock Changemamba: Remote sensing change detection with spatio-temporal state space model, 2024{\natexlab{a}}.

\bibitem[Chen et~al.(2024{\natexlab{b}})Chen, Chen, Liu, Li, Zou, and Shi]{chen2024rsmamba}
Keyan Chen, Bowen Chen, Chenyang Liu, Wenyuan Li, Zhengxia Zou, and Zhenwei Shi.
\newblock Rsmamba: Remote sensing image classification with state space model, 2024{\natexlab{b}}.

\bibitem[Chen et~al.(2022)Chen, Zhao, Zhang, Duan, Qi, and Zhao]{XiChen2022FocalClickTP}
Xi Chen, Zhiyan Zhao, Yilei Zhang, Manni Duan, Donglian Qi, and Hengshuang Zhao.
\newblock Focalclick: towards practical interactive image segmentation.
\newblock pages 1300--1309, 2022.

\bibitem[Cheng et~al.(2023)Cheng, Ye, Deng, Chen, Li, Wang, Su, Huang, Chen, Jiang, et~al.]{cheng2023sam}
Junlong Cheng, Jin Ye, Zhongying Deng, Jianpin Chen, Tianbin Li, Haoyu Wang, Yanzhou Su, Ziyan Huang, Jilong Chen, Lei Jiang, et~al.
\newblock Sam-med2d.
\newblock \emph{arXiv preprint arXiv:2308.16184}, 2023.

\bibitem[Chi et~al.(2022)Chi, Fan, Ramadge, and Rudnicky]{chi2022kerple}
Ta-Chung Chi, Ting-Han Fan, Peter~J Ramadge, and Alexander Rudnicky.
\newblock Kerple: Kernelized relative positional embedding for length extrapolation.
\newblock \emph{NeurIPS}, 35, 2022.

\bibitem[Chu et~al.(2021)Chu, Tian, Wang, Zhang, Ren, Wei, Xia, and Shen]{Chu_Tian_Wang_Zhang_Ren_Wei_Xia_Shen_2021}
Xiangxiang Chu, Zhi Tian, Yuqing Wang, Bo Zhang, Haibing Ren, Xiaolin Wei, Huaxia Xia, and Chunhua Shen.
\newblock Twins: Revisiting the design of spatial attention in vision transformers.
\newblock \emph{Neural Information Processing Systems,Neural Information Processing Systems}, 2021.

\bibitem[Dao(2023)]{dao2023flashattention2}
Tri Dao.
\newblock Flash{A}ttention-2: Faster attention with better parallelism and work partitioning.
\newblock 2023.

\bibitem[Dao et~al.(2022)Dao, Fu, Ermon, Rudra, and R{\'e}]{dao2022flashattention}
Tri Dao, Daniel~Y. Fu, Stefano Ermon, Atri Rudra, and Christopher R{\'e}.
\newblock Flash{A}ttention: Fast and memory-efficient exact attention with {IO}-awareness.
\newblock In \emph{Advances in Neural Information Processing Systems}, 2022.

\bibitem[Dehghani et~al.(2023)Dehghani, Mustafa, Djolonga, Heek, Minderer, Caron, Steiner, Puigcerver, Geirhos, Alabdulmohsin, et~al.]{dehghani2023patch}
Mostafa Dehghani, Basil Mustafa, Josip Djolonga, Jonathan Heek, Matthias Minderer, Mathilde Caron, Andreas Steiner, Joan Puigcerver, Robert Geirhos, Ibrahim Alabdulmohsin, et~al.
\newblock Patch n'pack: Navit, a vision transformer for any aspect ratio and resolution.
\newblock \emph{arXiv preprint arXiv:2307.06304}, 2023.

\bibitem[Dong et~al.(2022)Dong, Bao, Chen, Zhang, Yu, Yuan, Chen, and Guo]{Dong_Bao_Chen_Zhang_Yu_Yuan_Chen_Guo_2022}
Xiaoyi Dong, Jianmin Bao, Dongdong Chen, Weiming Zhang, Nenghai Yu, Lu Yuan, Dong Chen, and Baining Guo.
\newblock Cswin transformer: A general vision transformer backbone with cross-shaped windows.
\newblock In \emph{2022 IEEE/CVF Conference on Computer Vision and Pattern Recognition (CVPR)}, 2022.

\bibitem[Dosovitskiy et~al.(2021{\natexlab{a}})Dosovitskiy, Beyer, Kolesnikov, Weissenborn, Zhai, Unterthiner, Dehghani, Minderer, Heigold, Gelly, Uszkoreit, and Houlsby]{AlexeyDosovitskiy2020AnII}
Alexey Dosovitskiy, Lucas Beyer, Alexander Kolesnikov, Dirk Weissenborn, Xiaohua Zhai, Thomas Unterthiner, Mostafa Dehghani, Matthias Minderer, Georg Heigold, Sylvain Gelly, Jakob Uszkoreit, and Neil Houlsby.
\newblock An image is worth 16x16 words: Transformers for image recognition at scale.
\newblock In \emph{9th International Conference on Learning Representations, {ICLR} 2021, Virtual Event, Austria, May 3-7, 2021}. OpenReview.net, 2021{\natexlab{a}}.

\bibitem[Dosovitskiy et~al.(2021{\natexlab{b}})Dosovitskiy, Beyer, Kolesnikov, Weissenborn, Zhai, Unterthiner, Dehghani, Minderer, Heigold, Gelly, Uszkoreit, and Houlsby]{dosovitskiy2021image}
Alexey Dosovitskiy, Lucas Beyer, Alexander Kolesnikov, Dirk Weissenborn, Xiaohua Zhai, Thomas Unterthiner, Mostafa Dehghani, Matthias Minderer, Georg Heigold, Sylvain Gelly, Jakob Uszkoreit, and Neil Houlsby.
\newblock An image is worth 16x16 words: Transformers for image recognition at scale, 2021{\natexlab{b}}.

\bibitem[Fang et~al.(2023)Fang, Wang, Xie, Sun, Wu, Wang, Huang, Wang, and Cao]{fang2023eva}
Yuxin Fang, Wen Wang, Binhui Xie, Quan Sun, Ledell Wu, Xinggang Wang, Tiejun Huang, Xinlong Wang, and Yue Cao.
\newblock Eva: Exploring the limits of masked visual representation learning at scale.
\newblock In \emph{Proceedings of the IEEE/CVF Conference on Computer Vision and Pattern Recognition}, pages 19358--19369, 2023.

\bibitem[Fu et~al.(2024)Fu, Li, Cai, Wang, Wang, Shen, and Yao]{fu2024mddose}
Linjie Fu, Xia Li, Xiuding Cai, Yingkai Wang, Xueyao Wang, Yali Shen, and Yu Yao.
\newblock Md-dose: A diffusion model based on the mamba for radiotherapy dose prediction, 2024.

\bibitem[Gong et~al.(2021)Gong, Wang, Li, Chen, Yan, Tian, Chandra, et~al.]{18}
Chengyue Gong, Dilin Wang, Meng Li, Xinlei Chen, Zhicheng Yan, Yuandong Tian, Vikas Chandra, et~al.
\newblock Nasvit: Neural architecture search for efficient vision transformers with gradient conflict aware supernet training.
\newblock In \emph{ICLR}, 2021.

\bibitem[Gu and Dao(2023)]{gu2023mamba}
Albert Gu and Tri Dao.
\newblock Mamba: Linear-time sequence modeling with selective state spaces.
\newblock \emph{arXiv preprint arXiv:2312.00752}, 2023.

\bibitem[Gu et~al.(2022)Gu, Kwon, Wang, Ye, Li, Chen, Lai, Chandra, and Pan]{gu2022multi}
Jiaqi Gu, Hyoukjun Kwon, Dilin Wang, Wei Ye, Meng Li, Yu-Hsin Chen, Liangzhen Lai, Vikas Chandra, and David~Z Pan.
\newblock Multi-scale high-resolution vision transformer for semantic segmentation.
\newblock In \emph{Proceedings of the IEEE/CVF Conference on Computer Vision and Pattern Recognition}, pages 12094--12103, 2022.

\bibitem[Guo et~al.(2019)Guo, Qiu, Liu, Shao, Xue, and Zhang]{Guo_Qiu_Liu_Shao_Xue_Zhang_2019}
Qipeng Guo, Xipeng Qiu, Pengfei Liu, Yunfan Shao, Xiangyang Xue, and Zheng Zhang.
\newblock Star-transformer.
\newblock In \emph{Proceedings of the 2019 Conference of the North}, 2019.

\bibitem[Gupta et~al.(2019)Gupta, Doll{\'{a}}r, and Girshick]{AgrimGupta2019LVISAD}
Agrim Gupta, Piotr Doll{\'{a}}r, and Ross~B. Girshick.
\newblock {LVIS:} {A} dataset for large vocabulary instance segmentation.
\newblock In \emph{{IEEE} Conference on Computer Vision and Pattern Recognition, {CVPR} 2019, Long Beach, CA, USA, June 16-20, 2019}, pages 5356--5364. Computer Vision Foundation / {IEEE}, 2019.

\bibitem[Han et~al.(2023)Han, Pan, Han, Song, and Huang]{han2023flatten}
Dongchen Han, Xuran Pan, Yizeng Han, Shiji Song, and Gao Huang.
\newblock Flatten transformer: Vision transformer using focused linear attention.
\newblock In \emph{Proceedings of the IEEE/CVF International Conference on Computer Vision}, pages 5961--5971, 2023.

\bibitem[He et~al.(2021)He, Chen, Xie, Li, Doll{\'a}r, and Girshick]{KaimingHe2021MaskedAA}
Kaiming He, Xinlei Chen, Saining Xie, Yanghao Li, Piotr Doll{\'a}r, and Ross Girshick.
\newblock Masked autoencoders are scalable vision learners.
\newblock \emph{arXiv: Computer Vision and Pattern Recognition}, 2021.

\bibitem[He et~al.(2022)He, Chen, Xie, Li, Dollar, and Girshick]{20}
Kaiming He, Xinlei Chen, Saining Xie, Yanghao Li, Piotr Dollar, and Ross Girshick.
\newblock Masked autoencoders are scalable vision learners.
\newblock In \emph{CVPR}, 2022.

\bibitem[Ho et~al.(2019)Ho, Kalchbrenner, Weissenborn, and Salimans]{Ho_Kalchbrenner_Weissenborn_Salimans_2019}
Jonathan Ho, Nal Kalchbrenner, Dirk Weissenborn, and Tim Salimans.
\newblock Axial attention in multidimensional transformers.
\newblock \emph{Cornell University - arXiv,Cornell University - arXiv}, 2019.

\bibitem[Hu et~al.(2024)Hu, Baumann, Gui, Grebenkova, Ma, Fischer, and Ommer]{hu2024zigma}
Vincent~Tao Hu, Stefan~Andreas Baumann, Ming Gui, Olga Grebenkova, Pingchuan Ma, Johannes Fischer, and Björn Ommer.
\newblock Zigma: A dit-style zigzag mamba diffusion model, 2024.

\bibitem[Huang et~al.(2024{\natexlab{a}})Huang, Pei, You, Wang, Qian, and Xu]{huang2024localmamba}
Tao Huang, Xiaohuan Pei, Shan You, Fei Wang, Chen Qian, and Chang Xu.
\newblock Localmamba: Visual state space model with windowed selective scan, 2024{\natexlab{a}}.

\bibitem[Huang et~al.(2023)Huang, Yang, Sun, Zhang, Cao, Jiang, and Ji]{huang2023interformer}
You Huang, Hao Yang, Ke Sun, Shengchuan Zhang, Liujuan Cao, Guannan Jiang, and Rongrong Ji.
\newblock Interformer: Real-time interactive image segmentation.
\newblock In \emph{Proceedings of the IEEE/CVF International Conference on Computer Vision}, pages 22301--22311, 2023.

\bibitem[Huang et~al.(2024{\natexlab{b}})Huang, Lan, Cao, Lin, Zhang, Jiang, and Ji]{huang2024focsam}
You Huang, Zongyu Lan, Liujuan Cao, Xianming Lin, Shengchuan Zhang, Guannan Jiang, and Rongrong Ji.
\newblock Focsam: Delving deeply into focused objects in segmenting anything.
\newblock In \emph{Proceedings of the IEEE/CVF Conference on Computer Vision and Pattern Recognition}, pages 3120--3130, 2024{\natexlab{b}}.

\bibitem[Huang et~al.(2020)Huang, Wang, Wei, Huang, Shi, Liu, and Huang]{Huang_Wang_Wei_Huang_Shi_Liu_Huang_2020}
Zilong Huang, Xinggang Wang, Yunchao Wei, Lichao Huang, Humphrey Shi, Wenyu Liu, and Thomas~S. Huang.
\newblock Ccnet: Criss-cross attention for semantic segmentation.
\newblock \emph{IEEE Transactions on Pattern Analysis and Machine Intelligence}, page 1–1, 2020.

\bibitem[Jang and Kim(2019)]{JangWonDong2019InteractiveIS}
Won{-}Dong Jang and Chang{-}Su Kim.
\newblock Interactive image segmentation via backpropagating refinement scheme.
\newblock In \emph{{IEEE} Conference on Computer Vision and Pattern Recognition, {CVPR} 2019, Long Beach, CA, USA, June 16-20, 2019}, pages 5297--5306. Computer Vision Foundation / {IEEE}, 2019.

\bibitem[Ke et~al.(2023)Ke, Ye, Danelljan, Liu, Tai, Tang, and Yu]{sam_hq}
Lei Ke, Mingqiao Ye, Martin Danelljan, Yifan Liu, Yu-Wing Tai, Chi-Keung Tang, and Fisher Yu.
\newblock Segment anything in high quality.
\newblock In \emph{NeurIPS}, 2023.

\bibitem[Khan et~al.(2022)Khan, Naseer, Hayat, Zamir, Khan, and Shah]{khan2022transformers}
Salman Khan, Muzammal Naseer, Munawar Hayat, Syed~Waqas Zamir, Fahad~Shahbaz Khan, and Mubarak Shah.
\newblock Transformers in vision: A survey.
\newblock \emph{ACM computing surveys (CSUR)}, 54\penalty0 (10s):\penalty0 1--41, 2022.

\bibitem[Kirillov et~al.(2023)Kirillov, Mintun, Ravi, Mao, Rolland, Gustafson, Xiao, Whitehead, Berg, Lo, Doll{\'a}r, and Girshick]{kirillov2023segany}
Alexander Kirillov, Eric Mintun, Nikhila Ravi, Hanzi Mao, Chloe Rolland, Laura Gustafson, Tete Xiao, Spencer Whitehead, Alexander~C. Berg, Wan-Yen Lo, Piotr Doll{\'a}r, and Ross Girshick.
\newblock Segment anything.
\newblock \emph{arXiv:2304.02643}, 2023.

\bibitem[Konstantin~Sofiiuk(2021)]{KonstantinSofiiuk2021RevivingIT}
Anton~Konushin Konstantin~Sofiiuk, Ilia A.~Petrov.
\newblock Reviving iterative training with mask guidance for interactive segmentation.
\newblock \emph{arXiv: Computer Vision and Pattern Recognition}, 2021.

\bibitem[Kusupati et~al.(2022)Kusupati, Bhatt, Rege, Wallingford, Sinha, Ramanujan, Howard-Snyder, Chen, Kakade, Jain, and Farhadi]{31_felex}
Aditya Kusupati, Gantavya Bhatt, Aniket Rege, Matthew Wallingford, Aditya Sinha, Vivek Ramanujan, William Howard-Snyder, Kaifeng Chen, Sham Kakade, Prateek Jain, and Ali Farhadi.
\newblock Matryoshka representations for adaptive deployment.
\newblock \emph{arXiv preprint arXiv:2205.13147}, 2022.

\bibitem[Lai et~al.(2023)Lai, Tian, Chen, Li, Yuan, Liu, and Jia]{lai2023lisa}
Xin Lai, Zhuotao Tian, Yukang Chen, Yanwei Li, Yuhui Yuan, Shu Liu, and Jiaya Jia.
\newblock Lisa: Reasoning segmentation via large language model.
\newblock \emph{arXiv preprint arXiv:2308.00692}, 2023.

\bibitem[Lee et~al.(2023)Lee, Joshi, Turc, Hu, Liu, Eisenschlos, Khandelwal, Shaw, Chang, and Toutanova]{lee2023pix2struct}
Kenton Lee, Mandar Joshi, Iulia~Raluca Turc, Hexiang Hu, Fangyu Liu, Julian~Martin Eisenschlos, Urvashi Khandelwal, Peter Shaw, Ming-Wei Chang, and Kristina Toutanova.
\newblock Pix2struct: Screenshot parsing as pretraining for visual language understanding.
\newblock In \emph{ICML}, 2023.

\bibitem[Leroy et~al.(2023)Leroy, Revaud, Lucas, and Weinzaepfel]{leroy2023win}
Vincent Leroy, Jerome Revaud, Thomas Lucas, and Philippe Weinzaepfel.
\newblock Win-win: Training high-resolution vision transformers from two windows.
\newblock \emph{arXiv preprint arXiv:2310.00632}, 2023.

\bibitem[Li et~al.(2023{\natexlab{a}})Li, Vosselman, and Yang]{li2023interactive}
Kun Li, George Vosselman, and Michael~Ying Yang.
\newblock Interactive image segmentation with cross-modality vision transformers.
\newblock In \emph{Proceedings of the IEEE/CVF International Conference on Computer Vision}, pages 762--772, 2023{\natexlab{a}}.

\bibitem[Li et~al.(2023{\natexlab{b}})Li, Zhao, Wang, Cheng, Jin, Ji, Yuan, Liu, and Chen]{li2023multi}
Kehan Li, Yian Zhao, Zhennan Wang, Zesen Cheng, Peng Jin, Xiangyang Ji, Li Yuan, Chang Liu, and Jie Chen.
\newblock Multi-granularity interaction simulation for unsupervised interactive segmentation.
\newblock In \emph{Proceedings of the IEEE/CVF International Conference on Computer Vision}, pages 666--676, 2023{\natexlab{b}}.

\bibitem[Li et~al.(2024{\natexlab{a}})Li, Singh, and Grover]{li2024mamband}
Shufan Li, Harkanwar Singh, and Aditya Grover.
\newblock Mamba-nd: Selective state space modeling for multi-dimensional data, 2024{\natexlab{a}}.

\bibitem[Li et~al.(2024{\natexlab{b}})Li, Hong, and Fan]{li2024spikemba}
Wenrui Li, Xiaopeng Hong, and Xiaopeng Fan.
\newblock Spikemba: Multi-modal spiking saliency mamba for temporal video grounding, 2024{\natexlab{b}}.

\bibitem[Li et~al.()Li, Mao, Girshick, and He]{YanghaoLiExploringPV}
Yanghao Li, Hanzi Mao, Ross Girshick, and Kaiming He.
\newblock Exploring plain vision transformer backbones for object detection.

\bibitem[Li et~al.(2018)Li, Chen, and Koltun]{ZhuwenLi2018InteractiveIS}
Zhuwen Li, Qifeng Chen, and Vladlen Koltun.
\newblock Interactive image segmentation with latent diversity.
\newblock In \emph{2018 {IEEE} Conference on Computer Vision and Pattern Recognition, {CVPR} 2018, Salt Lake City, UT, USA, June 18-22, 2018}, pages 577--585. {IEEE} Computer Society, 2018.

\bibitem[Lin et~al.(2022{\natexlab{a}})Lin, Chen, Zhang, Li, Shen, Shen, and Ji]{34_flex}
Mingbao Lin, Mengzhao Chen, Yuxin Zhang, Ke Li, Yunhang Shen, Chunhua Shen, and Rongrong Ji.
\newblock Super vision transformer.
\newblock \emph{IJCV}, 2022{\natexlab{a}}.

\bibitem[Lin et~al.(2014)Lin, Maire, Belongie, Hays, Perona, Ramanan, Doll{\'a}r, and Zitnick]{TsungYiLin2014MicrosoftCC}
Tsung-Yi Lin, Michael Maire, Serge Belongie, James Hays, Pietro Perona, Deva Ramanan, Piotr Doll{\'a}r, and C.~Lawrence Zitnick.
\newblock Microsoft coco: Common objects in context.
\newblock \emph{Lecture Notes in Computer Science}, 2014.

\bibitem[Lin et~al.(2020)Lin, Zhang, Chen, Cheng, and Lu]{ZhengLin2020InteractiveIS}
Zheng Lin, Zhao Zhang, Lin{-}Zhuo Chen, Ming{-}Ming Cheng, and Shao{-}Ping Lu.
\newblock Interactive image segmentation with first click attention.
\newblock In \emph{2020 {IEEE/CVF} Conference on Computer Vision and Pattern Recognition, {CVPR} 2020, Seattle, WA, USA, June 13-19, 2020}, pages 13336--13345. {IEEE}, 2020.

\bibitem[Lin et~al.(2022{\natexlab{b}})Lin, Duan, Zhang, Guo, and Cheng]{ZhengLinFocusCutDI}
Zheng Lin, Zheng-Peng Duan, Zhao Zhang, Chun-Le Guo, and Ming-Ming Cheng.
\newblock Focuscut: Diving into a focus view in interactive segmentation.
\newblock pages 2637--2646, 2022{\natexlab{b}}.

\bibitem[Liu et~al.(2024{\natexlab{a}})Liu, Chen, Chen, Zhang, Zou, and Shi]{liu2024rscama}
Chenyang Liu, Keyan Chen, Bowen Chen, Haotian Zhang, Zhengxia Zou, and Zhenwei Shi.
\newblock Rscama: Remote sensing image change captioning with state space model, 2024{\natexlab{a}}.

\bibitem[Liu et~al.(2024{\natexlab{b}})Liu, Yang, Zhou, Xi, Yu, Yu, Liang, Shi, Zhang, Zheng, and Wang]{liu2024swinumamba}
Jiarun Liu, Hao Yang, Hong-Yu Zhou, Yan Xi, Lequan Yu, Yizhou Yu, Yong Liang, Guangming Shi, Shaoting Zhang, Hairong Zheng, and Shanshan Wang.
\newblock Swin-umamba: Mamba-based unet with imagenet-based pretraining, 2024{\natexlab{b}}.

\bibitem[Liu et~al.(2022{\natexlab{a}})Liu, Xu, Bertasius, and Niethammer]{QinLiu2022SimpleClickII}
Qin Liu, Zhenlin Xu, Gedas Bertasius, and Marc Niethammer.
\newblock Simpleclick: Interactive image segmentation with simple vision transformers.
\newblock \emph{arXiv preprint arXiv:2210.11006}, 2022{\natexlab{a}}.

\bibitem[Liu et~al.(2022{\natexlab{b}})Liu, Zheng, Planche, Karanam, Chen, Niethammer, and Wu]{QinLiu2022PseudoClickII}
Qin Liu, Meng Zheng, Benjamin Planche, Srikrishna Karanam, Terrence Chen, Marc Niethammer, and Ziyan Wu.
\newblock Pseudoclick: Interactive image segmentation with click imitation.
\newblock pages 728--745, 2022{\natexlab{b}}.

\bibitem[Liu et~al.(2024{\natexlab{c}})Liu, Cho, Bansal, and Niethammer]{liu2024rethinking}
Qin Liu, Jaemin Cho, Mohit Bansal, and Marc Niethammer.
\newblock Rethinking interactive image segmentation with low latency, high quality, and diverse prompts.
\newblock \emph{arXiv preprint arXiv:2404.00741}, 2024{\natexlab{c}}.

\bibitem[Liu et~al.(2024{\natexlab{d}})Liu, Tian, Zhao, Yu, Xie, Wang, Ye, and Liu]{liu2024vmamba}
Yue Liu, Yunjie Tian, Yuzhong Zhao, Hongtian Yu, Lingxi Xie, Yaowei Wang, Qixiang Ye, and Yunfan Liu.
\newblock Vmamba: Visual state space model.
\newblock \emph{arXiv preprint arXiv:2401.10166}, 2024{\natexlab{d}}.

\bibitem[Liu et~al.(2021)Liu, Lin, Cao, Hu, Wei, Zhang, Lin, and Guo]{liu2021swin}
Ze Liu, Yutong Lin, Yue Cao, Han Hu, Yixuan Wei, Zheng Zhang, Stephen Lin, and Baining Guo.
\newblock Swin transformer: Hierarchical vision transformer using shifted windows.
\newblock In \emph{Proceedings of the IEEE/CVF international conference on computer vision}, pages 10012--10022, 2021.

\bibitem[Liu et~al.(2022{\natexlab{c}})Liu, Hu, Lin, Yao, Xie, Wei, Ning, Cao, Zhang, Dong, et~al.]{liu2022swin}
Ze Liu, Han Hu, Yutong Lin, Zhuliang Yao, Zhenda Xie, Yixuan Wei, Jia Ning, Yue Cao, Zheng Zhang, Li Dong, et~al.
\newblock Swin transformer v2: Scaling up capacity and resolution.
\newblock In \emph{Proceedings of the IEEE/CVF conference on computer vision and pattern recognition}, pages 12009--12019, 2022{\natexlab{c}}.

\bibitem[Long et~al.(2024)Long, Zhou, Li, Lu, Ying, Luo, Ma, and Yan]{long2024dgmamba}
Shaocong Long, Qianyu Zhou, Xiangtai Li, Xuequan Lu, Chenhao Ying, Yuan Luo, Lizhuang Ma, and Shuicheng Yan.
\newblock Dgmamba: Domain generalization via generalized state space model, 2024.

\bibitem[Ma and Wang(2023)]{ma2023segment}
Jun Ma and Bo Wang.
\newblock Segment anything in medical images.
\newblock \emph{arXiv preprint arXiv:2304.12306}, 2023.

\bibitem[Ma et~al.(2024)Ma, Li, and Wang]{ma2024umamba}
Jun Ma, Feifei Li, and Bo Wang.
\newblock U-mamba: Enhancing long-range dependency for biomedical image segmentation, 2024.

\bibitem[Maninis et~al.(2018)Maninis, Caelles, Pont{-}Tuset, and Gool]{KevisKokitsiManinis2017DeepEC}
Kevis{-}Kokitsi Maninis, Sergi Caelles, Jordi Pont{-}Tuset, and Luc~Van Gool.
\newblock Deep extreme cut: From extreme points to object segmentation.
\newblock In \emph{2018 {IEEE} Conference on Computer Vision and Pattern Recognition, {CVPR} 2018, Salt Lake City, UT, USA, June 18-22, 2018}, pages 616--625. {IEEE} Computer Society, 2018.

\bibitem[Mazurowski et~al.(2023)Mazurowski, Dong, Gu, Yang, Konz, and Zhang]{mazurowski2023segment}
Maciej~A Mazurowski, Haoyu Dong, Hanxue Gu, Jichen Yang, Nicholas Konz, and Yixin Zhang.
\newblock Segment anything model for medical image analysis: an experimental study.
\newblock \emph{Medical Image Analysis}, 89:\penalty0 102918, 2023.

\bibitem[Patro and Agneeswaran(2024)]{patro2024simba}
Badri~N. Patro and Vijay~S. Agneeswaran.
\newblock Simba: Simplified mamba-based architecture for vision and multivariate time series, 2024.

\bibitem[Pei et~al.(2024)Pei, Huang, and Xu]{pei2024efficientvmamba}
Xiaohuan Pei, Tao Huang, and Chang Xu.
\newblock Efficientvmamba: Atrous selective scan for light weight visual mamba, 2024.

\bibitem[Perazzi et~al.(2016)Perazzi, Pont{-}Tuset, McWilliams, Gool, Gross, and Sorkine{-}Hornung]{FedericoPerazzi2016ABD}
Federico Perazzi, Jordi Pont{-}Tuset, Brian McWilliams, Luc~Van Gool, Markus~H. Gross, and Alexander Sorkine{-}Hornung.
\newblock A benchmark dataset and evaluation methodology for video object segmentation.
\newblock In \emph{2016 {IEEE} Conference on Computer Vision and Pattern Recognition, {CVPR} 2016, Las Vegas, NV, USA, June 27-30, 2016}, pages 724--732. {IEEE} Computer Society, 2016.

\bibitem[Press et~al.(2021)Press, Smith, Lewis, et~al.]{press2021train}
Ofir Press, Noah~A Smith, Mike Lewis, et~al.
\newblock Train short, test long: Attention with linear biases enables input length extrapolation.
\newblock \emph{arXiv preprint arXiv:2108.12409}, 2021.

\bibitem[Qiao et~al.(2024)Qiao, Yu, Guo, Chen, Zhao, Sun, Wu, and Liu]{qiao2024vlmamba}
Yanyuan Qiao, Zheng Yu, Longteng Guo, Sihan Chen, Zijia Zhao, Mingzhen Sun, Qi Wu, and Jing Liu.
\newblock Vl-mamba: Exploring state space models for multimodal learning, 2024.

\bibitem[Rabe and Staats(2021)]{DBLP:journals/corr/abs-2112-05682}
Markus~N. Rabe and Charles Staats.
\newblock Self-attention does not need o(n\({}^{\mbox{2}}\)) memory.
\newblock \emph{CoRR}, abs/2112.05682, 2021.

\bibitem[Rana et~al.(2023)Rana, Mahadevan, Hermans, and Leibe]{rana2023dynamite}
Amit~Kumar Rana, Sabarinath Mahadevan, Alexander Hermans, and Bastian Leibe.
\newblock Dynamite: Dynamic query bootstrapping for multi-object interactive segmentation transformer.
\newblock In \emph{Proceedings of the IEEE/CVF International Conference on Computer Vision}, pages 1043--1052, 2023.

\bibitem[Ren et~al.()Ren, Li, Wang, Xiao, and Chang]{Ren_Li_Wang_Xiao_Chang}
Pengzhen Ren, Changlin Li, Guangrun Wang, Yun Xiao, and QingDuXiaodanLiangXiaojun Chang.
\newblock Beyond fixation: Dynamic window visual transformer.

\bibitem[Research(2023)]{xformers_ops}
Facebook Research.
\newblock Xformers: Efficient attention operations, 2023.
\newblock Accessed: 2023-11-07.

\bibitem[Ruan and Xiang(2024)]{ruan2024vmunet}
Jiacheng Ruan and Suncheng Xiang.
\newblock Vm-unet: Vision mamba unet for medical image segmentation, 2024.

\bibitem[Shen et~al.(2024)Shen, Yi, Wu, Zhou, Zhang, Yan, and Wang]{shen2024gamba}
Qiuhong Shen, Xuanyu Yi, Zike Wu, Pan Zhou, Hanwang Zhang, Shuicheng Yan, and Xinchao Wang.
\newblock Gamba: Marry gaussian splatting with mamba for single view 3d reconstruction, 2024.

\bibitem[Shi et~al.(2024)Shi, Wu, Mao, Wang, and Darrell]{shi2024we}
Baifeng Shi, Ziyang Wu, Maolin Mao, Xin Wang, and Trevor Darrell.
\newblock When do we not need larger vision models?
\newblock \emph{arXiv preprint arXiv:2403.13043}, 2024.

\bibitem[Sofiiuk et~al.(2020)Sofiiuk, Petrov, Barinova, and Konushin]{KonstantinSofiiuk2020fBRSRB}
Konstantin Sofiiuk, Ilia~A. Petrov, Olga Barinova, and Anton Konushin.
\newblock {F-BRS:} rethinking backpropagating refinement for interactive segmentation.
\newblock In \emph{2020 {IEEE/CVF} Conference on Computer Vision and Pattern Recognition, {CVPR} 2020, Seattle, WA, USA, June 13-19, 2020}, pages 8620--8629. {IEEE}, 2020.

\bibitem[Song et~al.(2024)Song, Zhou, Li, Fan, Lu, and Ma]{song2024ba}
Yiran Song, Qianyu Zhou, Xiangtai Li, Deng-Ping Fan, Xuequan Lu, and Lizhuang Ma.
\newblock Ba-sam: Scalable bias-mode attention mask for segment anything model.
\newblock \emph{arXiv preprint arXiv:2401.02317}, 2024.

\bibitem[Strudel et~al.(2021)Strudel, Garcia, Laptev, and Schmid]{RobinStrudel2021SegmenterTF}
Robin Strudel, Ricardo Garcia, Ivan Laptev, and Cordelia Schmid.
\newblock Segmenter: Transformer for semantic segmentation.
\newblock pages 7262--7272, 2021.

\bibitem[Su et~al.(2024)Su, Ahmed, Lu, Pan, Bo, and Liu]{su2024roformer}
Jianlin Su, Murtadha Ahmed, Yu Lu, Shengfeng Pan, Wen Bo, and Yunfeng Liu.
\newblock Roformer: Enhanced transformer with rotary position embedding.
\newblock \emph{Neurocomputing}, 568:\penalty0 127063, 2024.

\bibitem[Touvron et~al.(2023)Touvron, Lavril, Izacard, Martinet, Lachaux, Lacroix, Rozi{\`e}re, Goyal, Hambro, Azhar, et~al.]{touvron2023llama}
Hugo Touvron, Thibaut Lavril, Gautier Izacard, Xavier Martinet, Marie-Anne Lachaux, Timoth{\'e}e Lacroix, Baptiste Rozi{\`e}re, Naman Goyal, Eric Hambro, Faisal Azhar, et~al.
\newblock Llama: Open and efficient foundation language models.
\newblock \emph{arXiv preprint arXiv:2302.13971}, 2023.

\bibitem[Vaswani et~al.(2017)Vaswani, Shazeer, Parmar, Uszkoreit, Jones, Gomez, Kaiser, and Polosukhin]{AshishVaswani2017AttentionIA}
Ashish Vaswani, Noam Shazeer, Niki Parmar, Jakob Uszkoreit, Llion Jones, Aidan~N. Gomez, Lukasz Kaiser, and Illia Polosukhin.
\newblock Attention is all you need.
\newblock In \emph{Advances in Neural Information Processing Systems 30: Annual Conference on Neural Information Processing Systems 2017, December 4-9, 2017, Long Beach, CA, {USA}}, pages 5998--6008, 2017.

\bibitem[Vaswani et~al.(2021)Vaswani, Ramachandran, Srinivas, Parmar, Hechtman, and Shlens]{Vaswani_Ramachandran_Srinivas_Parmar_Hechtman_Shlens_2021}
Ashish Vaswani, Prajit Ramachandran, Aravind Srinivas, Niki Parmar, Blake Hechtman, and Jonathon Shlens.
\newblock Scaling local self-attention for parameter efficient visual backbones.
\newblock In \emph{2021 IEEE/CVF Conference on Computer Vision and Pattern Recognition (CVPR)}, 2021.

\bibitem[Wan et~al.(2024)Wan, Wang, Yong, Zhang, Stepputtis, Sycara, and Xie]{wan2024sigma}
Zifu Wan, Yuhao Wang, Silong Yong, Pingping Zhang, Simon Stepputtis, Katia Sycara, and Yaqi Xie.
\newblock Sigma: Siamese mamba network for multi-modal semantic segmentation, 2024.

\bibitem[Wang et~al.(2023{\natexlab{a}})Wang, Liu, Zhao, Wu, Ma, Yu, Dai, Yang, Liu, Zhang, et~al.]{wang2023review}
Jiaqi Wang, Zhengliang Liu, Lin Zhao, Zihao Wu, Chong Ma, Sigang Yu, Haixing Dai, Qiushi Yang, Yiheng Liu, Songyao Zhang, et~al.
\newblock Review of large vision models and visual prompt engineering.
\newblock \emph{arXiv preprint arXiv:2307.00855}, 2023{\natexlab{a}}.

\bibitem[Wang et~al.(2021)Wang, Yao, Chen, Cai, He, and Liu]{Wang_Yao_Chen_Cai_He_Liu_2021}
Wenxiao Wang, Lu Yao, Long Chen, Deng Cai, Xiaofei He, and Wei Liu.
\newblock Crossformer: A versatile vision transformer based on cross-scale attention.
\newblock \emph{arXiv: Computer Vision and Pattern Recognition,arXiv: Computer Vision and Pattern Recognition}, 2021.

\bibitem[Wang et~al.(2023{\natexlab{b}})Wang, Dai, Chen, Huang, Li, Zhu, Hu, Lu, Lu, Li, et~al.]{wang2023internimage}
Wenhai Wang, Jifeng Dai, Zhe Chen, Zhenhang Huang, Zhiqi Li, Xizhou Zhu, Xiaowei Hu, Tong Lu, Lewei Lu, Hongsheng Li, et~al.
\newblock Internimage: Exploring large-scale vision foundation models with deformable convolutions.
\newblock In \emph{Proceedings of the IEEE/CVF Conference on Computer Vision and Pattern Recognition}, pages 14408--14419, 2023{\natexlab{b}}.

\bibitem[Wang et~al.(2024)Wang, Zheng, Zhang, Cui, and Li]{wang2024mambaunet}
Ziyang Wang, Jian-Qing Zheng, Yichi Zhang, Ge Cui, and Lei Li.
\newblock Mamba-unet: Unet-like pure visual mamba for medical image segmentation, 2024.

\bibitem[Wu et~al.(2023)Wu, Fu, Fang, Liu, Wang, Xu, Jin, and Arbel]{wu2023medical}
Junde Wu, Rao Fu, Huihui Fang, Yuanpei Liu, Zhaowei Wang, Yanwu Xu, Yueming Jin, and Tal Arbel.
\newblock Medical sam adapter: Adapting segment anything model for medical image segmentation.
\newblock \emph{arXiv preprint arXiv:2304.12620}, 2023.

\bibitem[Xie et~al.(2021)Xie, Wang, Yu, Anandkumar, Alvarez, and Luo]{Xie_Wang_Yu_Anandkumar_Alvarez_Luo_2021}
Enze Xie, Wenhai Wang, Zhiding Yu, Animashree Anandkumar, JoseM. Alvarez, and Ping Luo.
\newblock Segformer: Simple and efficient design for semantic segmentation with transformers.
\newblock \emph{Cornell University - arXiv,Cornell University - arXiv}, 2021.

\bibitem[Xiong et~al.(2023)Xiong, Varadarajan, Wu, Xiang, Xiao, Zhu, Dai, Wang, Sun, Iandola, et~al.]{xiong2023efficientsam}
Yunyang Xiong, Bala Varadarajan, Lemeng Wu, Xiaoyu Xiang, Fanyi Xiao, Chenchen Zhu, Xiaoliang Dai, Dilin Wang, Fei Sun, Forrest Iandola, et~al.
\newblock Efficientsam: Leveraged masked image pretraining for efficient segment anything.
\newblock \emph{arXiv preprint arXiv:2312.00863}, 2023.

\bibitem[Xu et~al.(2024{\natexlab{a}})Xu, Chen, Huang, Wu, and Lai]{xu2024structured}
Long Xu, Yongquan Chen, Rui Huang, Feng Wu, and Shiwu Lai.
\newblock Structured click control in transformer-based interactive segmentation.
\newblock \emph{arXiv preprint arXiv:2405.04009}, 2024{\natexlab{a}}.

\bibitem[Xu et~al.(2016)Xu, Price, Cohen, Yang, and Huang]{NingXu2016DeepIO}
Ning Xu, Brian~L. Price, Scott Cohen, Jimei Yang, and Thomas~S. Huang.
\newblock Deep interactive object selection.
\newblock In \emph{2016 {IEEE} Conference on Computer Vision and Pattern Recognition, {CVPR} 2016, Las Vegas, NV, USA, June 27-30, 2016}, pages 373--381. {IEEE} Computer Society, 2016.

\bibitem[Xu et~al.(2024{\natexlab{b}})Xu, Yuan, Shi, Qi, Wang, Yang, Li, Chen, Tong, Ghanem, et~al.]{xu2024rap}
Shilin Xu, Haobo Yuan, Qingyu Shi, Lu Qi, Jingbo Wang, Yibo Yang, Yining Li, Kai Chen, Yunhai Tong, Bernard Ghanem, et~al.
\newblock Rap-sam: Towards real-time all-purpose segment anything.
\newblock \emph{arXiv preprint arXiv:2401.10228}, 2024{\natexlab{b}}.

\bibitem[Yan et~al.(2023)Yan, Wang, Liu, Jiang, Hu, Tang, Kang, and Gavves]{yan2023piclick}
Cilin Yan, Haochen Wang, Jie Liu, Xiaolong Jiang, Yao Hu, Xu Tang, Guoliang Kang, and Efstratios Gavves.
\newblock Piclick: Picking the desired mask in click-based interactive segmentation.
\newblock \emph{arXiv preprint arXiv:2304.11609}, 2023.

\bibitem[Yang et~al.(2024{\natexlab{a}})Yang, Chen, Espinosa, Ericsson, Wang, Liu, and Crowley]{yang2024plainmamba}
Chenhongyi Yang, Zehui Chen, Miguel Espinosa, Linus Ericsson, Zhenyu Wang, Jiaming Liu, and Elliot~J. Crowley.
\newblock Plainmamba: Improving non-hierarchical mamba in visual recognition, 2024{\natexlab{a}}.

\bibitem[Yang et~al.()Yang, Li, Zhang, Dai, Xiao, Yuan, Gao, Redmond, Cloud, and Ai]{Yang_Li_Zhang_Dai_Xiao_Yuan_Gao_Redmond_Cloud_Ai}
Jianwei Yang, Chunyuan Li, Pengchuan Zhang, Xiyang Dai, Bin Xiao, Lu Yuan, Jianfeng Gao, MicrosoftResearchAt Redmond, Microsoft Cloud, and + Ai.
\newblock Focal self-attention for local-global interactions in vision transformers.

\bibitem[Yang et~al.(2024{\natexlab{b}})Yang, Ma, Yao, Zhong, Zhang, and Wang]{yang2024remamber}
Yuhuan Yang, Chaofan Ma, Jiangchao Yao, Zhun Zhong, Ya Zhang, and Yanfeng Wang.
\newblock Remamber: Referring image segmentation with mamba twister, 2024{\natexlab{b}}.

\bibitem[Ye et~al.(2024)Ye, Chen, Wang, Zhang, Li, and Zhang]{ye2024pmamba}
Zi Ye, Tianxiang Chen, Fangyijie Wang, Hanwei Zhang, Guanxi Li, and Lijun Zhang.
\newblock P-mamba: Marrying perona malik diffusion with mamba for efficient pediatric echocardiographic left ventricular segmentation, 2024.

\bibitem[Yin et~al.(2021)Yin, Vahdat, Alvarez, Mallya, Kautz, and Molchanov]{yin2021adavit}
Hongxu Yin, Arash Vahdat, Jose Alvarez, Arun Mallya, Jan Kautz, and Pavlo Molchanov.
\newblock Adavit: Adaptive tokens for efficient vision transformer.
\newblock \emph{arXiv preprint arXiv:2112.07658}, 2021.

\bibitem[Yu et~al.(2020)Yu, Jin, Liu, Bender, Kindermans, Tan, Huang, Song, Pang, and Le]{63}
Jiahui Yu, Pengchong Jin, Hanxiao Liu, Gabriel Bender, Pieter-Jan Kindermans, Mingxing Tan, Thomas Huang, Xiaodan Song, Ruoming Pang, and Quoc Le.
\newblock Bignas: Scaling up neural architecture search with big single-stage models.
\newblock In \emph{ECCV}, 2020.

\bibitem[Yu et~al.(2023)Yu, Feng, Feng, Liu, Jin, Zeng, and Chen]{yu2023inpaint}
Tao Yu, Runseng Feng, Ruoyu Feng, Jinming Liu, Xin Jin, Wenjun Zeng, and Zhibo Chen.
\newblock Inpaint anything: Segment anything meets image inpainting.
\newblock \emph{arXiv preprint arXiv:2304.06790}, 2023.

\bibitem[Yuan et~al.(2021)Yuan, Fu, Huang, Lin, Zhang, Chen, and Wang]{yuan2021hrformer}
Yuhui Yuan, Rao Fu, Lang Huang, Weihong Lin, Chao Zhang, Xilin Chen, and Jingdong Wang.
\newblock Hrformer: High-resolution vision transformer for dense predict.
\newblock \emph{Advances in Neural Information Processing Systems}, 34:\penalty0 7281--7293, 2021.

\bibitem[Zhang et~al.(2023)Zhang, Han, Qiao, Kim, Bae, Lee, and Hong]{zhang2023faster}
Chaoning Zhang, Dongshen Han, Yu Qiao, Jung~Uk Kim, Sung-Ho Bae, Seungkyu Lee, and Choong~Seon Hong.
\newblock Faster segment anything: Towards lightweight sam for mobile applications.
\newblock \emph{arXiv preprint arXiv:2306.14289}, 2023.

\bibitem[Zhang et~al.(2020)Zhang, Liew, Wei, Wei, and Zhao]{ShiyinZhang2020InteractiveOS}
Shiyin Zhang, Jun~Hao Liew, Yunchao Wei, Shikui Wei, and Yao Zhao.
\newblock Interactive object segmentation with inside-outside guidance.
\newblock In \emph{2020 {IEEE/CVF} Conference on Computer Vision and Pattern Recognition, {CVPR} 2020, Seattle, WA, USA, June 13-19, 2020}, pages 12231--12241. {IEEE}, 2020.

\bibitem[Zhang et~al.(2024)Zhang, Chen, Qu, Yuille, and Zhou]{zhang2024leveraging}
Tiezheng Zhang, Xiaoxi Chen, Chongyu Qu, Alan Yuille, and Zongwei Zhou.
\newblock Leveraging ai predicted and expert revised annotations in interactive segmentation: Continual tuning or full training?
\newblock \emph{arXiv preprint arXiv:2402.19423}, 2024.

\bibitem[Zhao et~al.(2024)Zhao, Zhang, Zhao, Ding, Huang, and Wang]{zhao2024cobra}
Han Zhao, Min Zhang, Wei Zhao, Pengxiang Ding, Siteng Huang, and Donglin Wang.
\newblock Cobra: Extending mamba to multi-modal large language model for efficient inference, 2024.

\bibitem[Zhou et~al.(2023)Zhou, Wang, Zhao, Li, Huang, Meng, and Zheng]{zhou2023interactive}
Minghao Zhou, Hong Wang, Qian Zhao, Yuexiang Li, Yawen Huang, Deyu Meng, and Yefeng Zheng.
\newblock Interactive segmentation as gaussion process classification.
\newblock In \emph{Proceedings of the IEEE/CVF Conference on Computer Vision and Pattern Recognition}, pages 19488--19497, 2023.

\bibitem[Zhu et~al.(2024)Zhu, Liao, Zhang, Wang, Liu, and Wang]{zhu2024vision}
Lianghui Zhu, Bencheng Liao, Qian Zhang, Xinlong Wang, Wenyu Liu, and Xinggang Wang.
\newblock Vision mamba: Efficient visual representation learning with bidirectional state space model.
\newblock \emph{arXiv preprint arXiv:2401.09417}, 2024.

\bibitem[Zhuoran et~al.(2021)Zhuoran, Mingyuan, Haiyu, Shuai, and Hongsheng]{Zhuoran_Mingyuan_Haiyu_Shuai_Hongsheng_2021}
Shen Zhuoran, Zhang Mingyuan, Zhao Haiyu, Yi Shuai, and Li Hongsheng.
\newblock Efficient attention: Attention with linear complexities.
\newblock In \emph{2021 IEEE Winter Conference on Applications of Computer Vision (WACV)}, 2021.

\end{thebibliography}
